%% file: main.tex
\pdfoutput=1
\documentclass[11pt]{article}
\usepackage{floatrow}
\usepackage{emnlp2021}

\usepackage{times}
\usepackage{latexsym}
\usepackage{xspace,mfirstuc,tabulary}
\usepackage[utf8]{inputenc}

\usepackage[ruled,vlined]{algorithm2e}
\usepackage{listings}

\usepackage{latexsym}
\usepackage{url}
\usepackage{paralist}
\usepackage{multicol,graphicx,mathrsfs,pifont,amscd,latexsym,color,url, pifont, subfigure,epstopdf,setspace,multirow,colortbl, bbm,listings, balance,color,indentfirst,enumitem, subfigure,amsmath,amssymb, verbatim,microtype}
\usepackage{graphicx,booktabs, paralist}

\input{mathcommand.tex}

\input{pseudocode_style.tex}

\usepackage[T1]{fontenc}
\definecolor{lb}{HTML}{E0ECF8}

\definecolor{db}{HTML}{537FCA}
\usepackage{caption}
\usepackage[utf8]{inputenc}
\usepackage{upgreek}
\usepackage{microtype}

\usepackage{amssymb}% http://ctan.org/pkg/amssymb
\usepackage{pifont}% http://ctan.org/pkg/pifont
\usepackage{changepage}

\usepackage{graphics}

\newcommand{\method}{O\textsc{reo}LM\xspace}
\newcommand{\KG}{\mathcal{KG}\xspace}

\newcommand{\rw}{CRW\xspace}

\usepackage[hybrid]{markdown}

\title{Empowering Language Models with Knowledge Graph Reasoning for Question Answering}

\author{Ziniu Hu$^1$, Yichong Xu$^2$, Wenhao Yu$^3$, Shuohang Wang$^2$\\ \textbf{Ziyi Yang$^2$, Chenguang Zhu$^2$, Kai-Wei Chang$^1$, Yizhou Sun$^1$}\\
  $^1$University of California, Los Angeles\\
  $^2$Microsoft Cognitive Services Research, $^3$University of Notre Dame
}

\begin{document}
\maketitle

\begin{abstract}
\input{sections/abstract}
\end{abstract}

\section{Introduction}\label{sec:introduction}\input{sections/intro_new}

% \section{Background and Preliminary}\label{sec:preliminary}\input{sections/preliminary}

\section{Methodology}\label{sec:methodology}\input{sections/methodology}

% \section{Pre-Training \method to Reason}\label{sec:pretrain}\input{sections/pretrain}

\section{Experiments}\label{sec:experiment}\input{sections/experiments}

\section{Related Work}\label{sec:related}\input{sections/relatedwork}

\section{Conclusion}\label{sec:conclusion}\input{sections/conclusion}

\section{Limitations}\label{sec:limitation}

\input{sections/limitation}

\bibliography{emnlp}
\bibliographystyle{acl_natbib}

\newpage
\appendix

\clearpage

\input{sections/appendix.tex}

\end{document}

%% file: mathcommand.tex
\newcommand{\rb}{\boldsymbol{\gamma}}

\newcommand{\pib}{\boldsymbol{\pi}}
\newcommand{\Rho}{\boldsymbol{\varrho}}

\newcommand{\Db}{\boldsymbol{D}}

\newcommand{\Ib}{\boldsymbol{I}}

\newcommand{\Pb}{\boldsymbol{P}}

%% file: pseudocode_style.tex
\definecolor{codegreen}{rgb}{0,0.6,0}
\definecolor{codegray}{rgb}{0.5,0.5,0.5}
\definecolor{codepurple}{rgb}{0.58,0,0.82}
\definecolor{backcolour}{rgb}{0.95,0.95,0.92}

\lstdefinestyle{style_full}{
    commentstyle=\color{codegreen},
    keywordstyle=\color{magenta},
    numberstyle=\tiny\color{codegray},
    stringstyle=\color{codepurple},
    basicstyle=\ttfamily\footnotesize,
    breakatwhitespace=false,         
    breaklines=true,                 
    captionpos=b,                    
    keepspaces=true,                 
    numbers=left,                    
    numbersep=8pt,                  
    showspaces=false,                
    showstringspaces=false,
    showtabs=false,                  
    tabsize=4,
    xleftmargin=3em
}

\lstdefinestyle{style_snippet}{  
    commentstyle=\color{codegreen},
    keywordstyle=\color{magenta},
    numberstyle=\tiny\color{codegray},
    stringstyle=\color{codepurple},
    basicstyle=\ttfamily\scriptsize,
    breakatwhitespace=false,         
    breaklines=true,                 
    captionpos=b,                    
    keepspaces=true,                 
    numbers=none,                    
    numbersep=0pt,                  
    showspaces=false,                
    showstringspaces=false,
    showtabs=false,                  
    tabsize=2,
    xleftmargin=-1.5em
}

\SetKwComment{Comment}{\color{green!50!black}// }{}

\SetKwProg{Function}{function}{}{}

%% file: sections/abstract.tex
% Existing open-domain QA systems require to be trained on a large human-annotated QA dataset. 

Answering open-domain questions requires world knowledge about in-context entities. As pre-trained Language Models (\texttt{LM}s) lack the power to store all required knowledge, external knowledge sources, such as knowledge graphs, are often used to augment \texttt{LM}s. 
%However, most existing methods do not allow \texttt{LM} to interact with $\mathcal{KG}$.
In this work, we propose kn\underline{O}wledge \underline{RE}as\underline{O}ning empowered \underline{L}anguage \underline{M}odel
(\method), which consists of a novel Knowledge Interaction Layer that can be flexibly plugged into existing Transformer-based \texttt{LM}s to interact with a differentiable Knowledge Graph Reasoning module collaboratively. In this way, \texttt{LM} guides KG to walk towards the desired answer, while the retrieved knowledge improves \texttt{LM}.
By adopting \method to RoBERTa and T5, we show significant performance gain, achieving state-of-art results in the \textit{Closed-Book} setting. The performance enhancement is mainly from the KG reasoning's capacity to infer missing relational facts. In addition, 
\method  provides reasoning paths as rationales to interpret the model's decision.
% \looseness=-1

%% file: sections/intro_new.tex
\noindent Open-Domain Question Answering (ODQA), one of the most knowledge-intensive NLP tasks, requires QA models to infer out-of-context knowledge to the given single question. 
Following the pioneering work by~\citet{DBLP:conf/acl/ChenFWB17}, ODQA systems often assume to access an external text corpus (e.g., Wikipedia) as an external knowledge source.
Due to the large scale of such textual knowledge sources (e.g., 20GB for Wikipedia), it cannot be encoded in the model parameters. Therefore, most works retrieve relevant passages as knowledge and thus named \textit{Open-Book} models~\citep{DBLP:conf/emnlp/RobertsRS20}, with an analogy of referring to textbooks during an exam. Another line of \textit{Closed-book} models~\citep{DBLP:conf/emnlp/RobertsRS20} assume knowledge could be stored implicitly in parameters of Language Models (\texttt{LM}, e.g. BERT~\citep{DBLP:conf/naacl/DevlinCLT19} and T5~\citep{DBLP:journals/jmlr/RaffelSRLNMZLL20}). These \texttt{LM}s directly generate answers without retrieving from an external corpus and thus benefit from faster inference speed and simpler training. However, current \texttt{LM}s still miss a large portion of factual knowledge~\citep{DBLP:conf/emnlp/PornerWS20, DBLP:conf/eacl/LewisSR21}, and are not competitive with \textit{Open-Book} models.

% To improve \textit{Closed-book} QA performance, a line of works seek more compressed knowledge representation than text corpus so that it could be stored in memory and directly enhance \texttt{LM} without an additional retrieval model. Knowledge Graph ($\KG$), which captures world knowledge explicitly via relational triplets between entities, is a natural choice for enhancing \texttt{LM}. There are a number of nice properties of $\KG$ as knowledge source: 1) it's a more abstract and compressed representation of knowledge than text corpus, and thus could be 

\begin{figure}[t!]
    \centering
    \includegraphics[width=1\columnwidth]{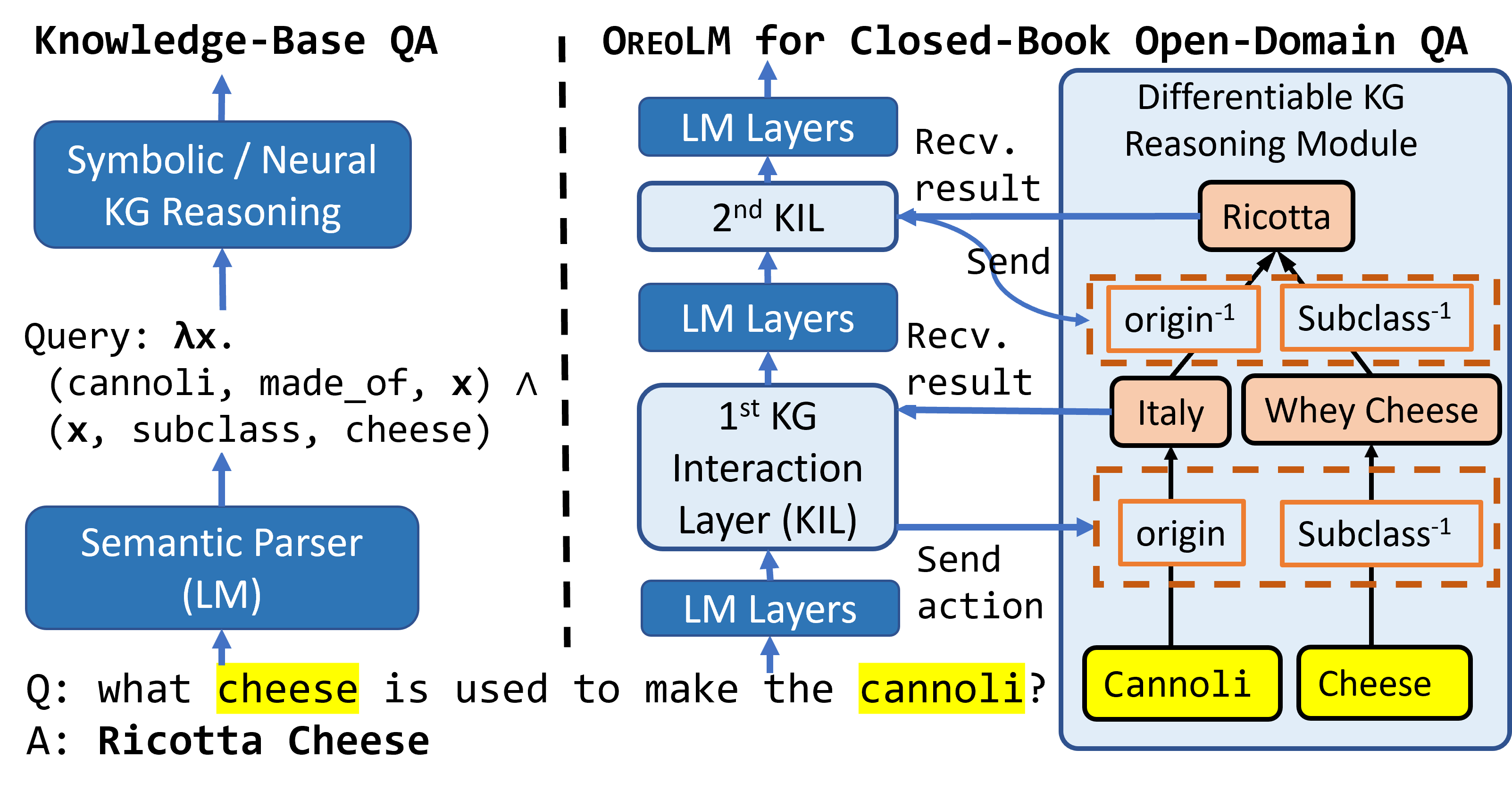}
    \vspace{-10pt}
    \caption{An Illustrative figure of \method. Compared with previous KBQA systems that stack reasoner on top of \texttt{LM}, \method enables interaction between the two. %\YS{use the same color blocks for cheese and cannoli in the question to make it clear those two entities in KG are from the question? }
    }
    \label{fig:intro}
\end{figure}

To improve the knowledge coverage of \texttt{LM}, one natural choice is to leverage knowledge stored in Knowledge Graph ($\KG$, e.g. FreeBase~\citep{DBLP:conf/sigmod/BollackerEPST08} and WikiData~\citep{DBLP:journals/cacm/VrandecicK14}), which explicitly encodes world knowledge via relational triplets between entities. There are several good properties of $\KG$:
%as a knowledge source: 
1) a $\KG$ triplet is a more abstract and compressed representation of knowledge than text, and thus $\KG$ could be stored in memory and directly enhance \texttt{LM} without using an additional retrieval model; 2) the structural nature of $\KG$ could support logical reasoning~\citep{DBLP:conf/iclr/RenHL20} and infer missing knowledge through high-order paths~\citep{DBLP:conf/emnlp/LaoMC11,DBLP:conf/iclr/DasDZVDKSM18}.
Taking the question ``what cheese is used to make the desert cannoli?'' as an example,   even if this relational fact is missing in $\KG$, we could still leverage high-order relationships, e.g., both Ricotta Cheese and Cannoli are specialties in Italy, to infer the answer ``Ricotta Cheese.''
%Such reasoning capability is essential for current \texttt{LM} that misses lots of knowledge. 

In light of the good properties of $\KG$, there are several efforts to build Knowledge Base Question Answering (KBQA) systems.
As is illustrated in Figure~\ref{fig:intro}(a), most KBQA models use \texttt{LM} as a parser to map textual questions into a structured form (e.g., SQL query or subgraph), and then based on $\KG$, the queries could be executed by
symbolic reasoning~\citep{DBLP:conf/emnlp/BerantCFL13} or neural reasoning (e.g. Graph Neural Networks) ~\citep{DBLP:conf/emnlp/SunBC19} to get the answer. 
Another recent line of research \cite{DBLP:conf/naacl/VergaSSC21, DBLP:journals/corr/abs-2010-00796} tries to encode the knowledge graph as the \emph{memory} into \texttt{LM} parameters. 
However, for most methods discussed above, \texttt{LM} is not interacting with $\KG$ to correctly understand the question, and the answer is usually restricted to a node or edge in $\KG$.
% In addition, most KBQA assume answer is an entity contained in $\KG$. 
% These limitations restrict existing KBQA systems for answering open-domain questions.

% and it remains an open question how to effectively integrate $\KG$ reasoning with \texttt{LM}.

% These methods require large amounts of annotated queries to train semantic parser. More recent methods propose adding knowledge graph embeddings~\citep{DBLP:conf/acl/ZhangHLJSL19,DBLP:conf/aaai/LiuW0PY21} or conducting $\KG$-guided pre-training~\citep{DBLP:journals/corr/abs-2010-00796, DBLP:conf/acl/KeJRCWSZH21, DBLP:journals/corr/abs-2010-00796}.

In this paper, we
propose kn\underline{O}wledge \underline{RE}as\underline{O}ning empowered \underline{L}anguage \underline{M}odel
(\method), a model architecture that can be applied to Transformer-based \texttt{LM}s to improve \textit{Closed-Book} ODQA. As is illustrated in Figure~\ref{fig:intro}(b), the key component is the Knowledge Interaction Layers (\texttt{KIL}) inserted amid \texttt{LM} layers, which is like cream filling within two waffles, leading to our model's name O\textsc{reo}. \texttt{KIL} interacts with a $\KG$ reasoning module, in which we maintain different reasoning paths for each entity in the question.
We formulate the retrieval and reasoning process as a contextualized \emph{random walk} over the $\KG$, starting from the in-context entities. Each \texttt{KIL} is responsible for one reasoning step. It first predicts a relation distribution for every in-context entity, and then the $\KG$ reasoning module traverses the graph following the predicted relation distribution. The reasoning result in each step is summarized as a weighted averaged embedding over the retrieved entities from the traversal.
% \looseness=-1
%current reasoning steps (nodes). 

% From bottom to top, each \texttt{KIL} first predict relation for every in-context entity, send the relation to $\KG$. The $\KG$ reasoning module then walks over the graph to expand the reasoning path and reach a new entity as retrieved knowledge. 
% Such a procedure is non-parametric, and we could keep track of each intermediate state to interpret the model decision. 
% \texttt{KIL} received the retrieved knowledge and injected it into \texttt{LM} so that it could better understand the question and make correct relation action at the next step.
By stacking $T$ layers of \texttt{KIL}, \method can retrieve entities that are $T$-hop away from in-context entities and help \texttt{LM} to answer open questions that require out-of-context knowledge or multi-hop reasoning. The whole procedure is fully differentiable, and thus \method learns and infers in an end-to-end manner. 
%In addition, the generated reasoning paths could serve as a rationale to interpret why \method generates a particular answer, which is similar to the parsed tree in traditional KBQA, but \method does not rely on them to get answers.
We further introduce how to pre-train \method over unlabelled Wikipedia corpus. In addition to the salient entity span masking objective, we introduce two self-supervised objectives %from weak supervision 
to guide \method to learn better entity and relation representations and how to reason over them.

We test \method with RoBERTa and T5 as our base \texttt{LM}s. By evaluating on several single-hop ODQA datasets in \textit{closed-book} setting, we show that \method outperforms existing baselines with fewer model parameters. 
Specifically, \method helps more for questions with missing relations in $\KG$, and questions that require multi-hop reasoning.
We further show that \method can serve as a backbone for \textit{open-book} setting and achieves comparable performance compared with the state-of-the-art QA systems with dedicated design. In addition, \method has better interpretability as it can generate reasoning paths for the answered question and summarize general relational rules to infer missing relations. 

% In addition, the design of \method allows it to accommodate new or modified relational facts, making the \method easier to debug and fast adapt to new knowledge.

This key contributions are as follows:
\begin{compactitem}
\item 
We propose \method to integrate symbolic knowledge graph reasoning with neural LMs. Different from prior works, \method can be seamlessly plugged into existing \texttt{LM}s. 
\item We pretrain \method with RoBERTa and T5 to on the Wikipedia corpus. \method can bring significant performance gain on ODQA.
% , including new state-of-art results in \textit{Closed-Book} settings.
\item \method offers interpretable reasoning paths for answering the question and high-order reasoning rules as rationales.
\end{compactitem}

%% file: sections/methodology.tex
\begin{figure*}[t!]
%\vspace{-.6in}
\hspace{-0.2in}
    \centering
    \includegraphics[width=1\columnwidth]{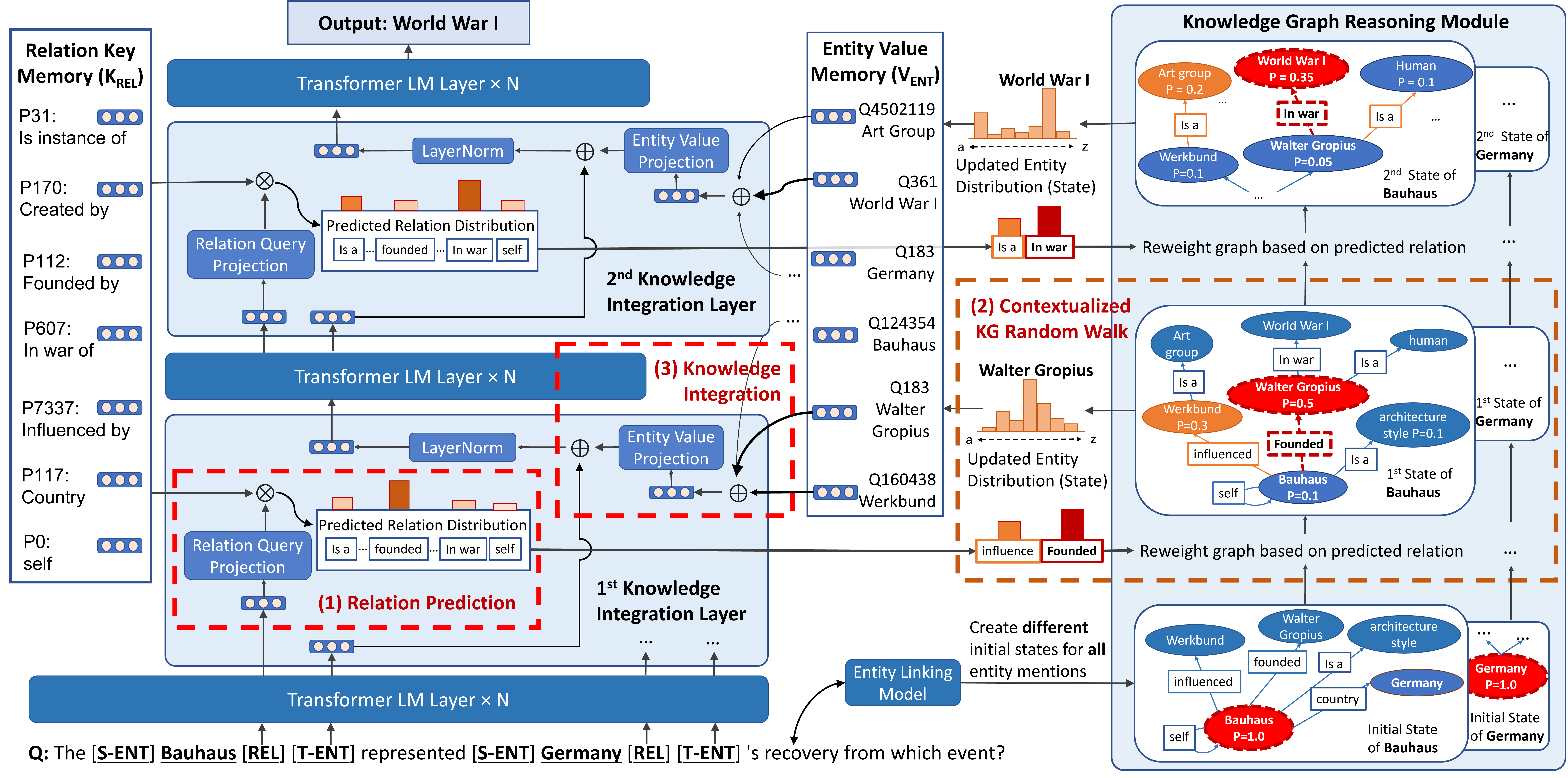}
    \caption{\textbf{Model architecture of \method}. Three key procedures are highlighted in red dotted box: 1) \textbf{Relation Prediction} (Sec.~\ref{sec:relation}): Knowledge Interaction Layers (\texttt{KIL}) predicts relation action for each entity mention. 2) \textbf{One-step State Transition} (Sec.~\ref{sec:crw}): Based on the predicted relation, $\KG$ re-weights each graph and conduct contextualized random walk to update entity distribution state. 3) \textbf{Knowledge Integration} (Sec.~\ref{sec:lm}): An weighted aggregated entity embedding is added into a placeholder token as retrieved knowledge. 
    % By repeating such procedure $T$ times, \method could retrieve knowledge that are $T$-hop away from initial entities in the question, and answer complex open-domain questions.
    }
    \label{fig:framework}
%  \vspace{-.1in}
\end{figure*}

\paragraph{Preliminary}
We denote a Knowledge Graph $\KG = \big(\mathcal{E}, \mathcal{R}, \mathcal{A} = \{A_r \}_{r \in \mathcal{R}} \big)$, where each $e \in \mathcal{E}$ and $r \in \mathcal{R}$ is entity node and relation label.
% as well as a binary function $r: \mathcal{E} \times \mathcal{E} \rightarrow \{\text{True}, \text{False}\}$ indicating whether relation $r$ holds between a pair of entities
$A_r \in \{0,1\}^{|\mathcal{E}| \times |\mathcal{E}|} $ is a sparse adjacency matrix indicating whether relation $r$ holds between a pair of entities.
% , s.t. $A_r[s,t] = 1 \Leftrightarrow  r(s,t) $.
The task of knowledge graph reasoning aims at answering a factoid query $(s, r, ?)$, i.e., which target entity has relation $r$ with the source entity $s$. If $\KG$ is complete, we could simply get answers by checking the adjacency matrix, i.e., $\{\forall t : A_r[s,t]=1\}$. For incomplete $\KG$ where many relational facts are missing, path-based reasoning approaches~\cite{DBLP:conf/emnlp/LaoMC11, DBLP:conf/emnlp/XiongHW17, DBLP:conf/iclr/DasDZVDKSM18} 
have been proposed to answer the one-hop query via finding multi-hop paths. For example, to answer the query $(s, \text{Mother}, ?)$, a path $s \xrightarrow{\text{Father}} j \xrightarrow{\text{Wife}} t$ could reach the target answer $t$. 
% \YS{say something about the connection of these preliminaries to our model.}
% Previous work mainly focus on reasoning over structured $\KG$ only, while in this paper we try to integrate $\KG$ into neural Language Models to solve 
In this paper we try to integrate symbolic $\KG$ reasoning into neural \texttt{LM}s and help it deal with ODQA problems.
% \looseness=-1

% \vspace{0.05in}
\paragraph{Overview of \method}
We illustrate the overall architecture of \method in Figure~\ref{fig:framework}. All the \colorbox{lb}{light blue blocks} are our added components to support $\KG$ reasoning, while the \colorbox{db}{dark blue} Transformer layers are knowledge-injected \texttt{LM}. The key component of \method for conducting $\KG$ reasoning is the Knowledge Interaction Layers (\texttt{KIL}), which are added amid \texttt{LM} layers to enable deeper interaction with the $\KG$. %by sending instruction to and receiving knowledge.

Given a question $q=$ ``The Bauhaus represented Germany's recovery from which event?'', QA model needs to extract knowledge about all $n$ in-context entity mentions $M=\{m_i\}_{i=1}^n$, e.g., the history of ``Germany'' at the time when ``Bauhaus'' is founded, to get the answer $a=$ ``World War I''. Such open-domain Q\&A can be abstracted as $P(a | q, M)$. 
% \looseness=-1
% In this paper, we propose to empower Language Models (\texttt{LM}, e.g., BERT and T5) with knowledge graph reasoning to retrieve necessary knowledge and answer such query in an end-to-end manner. Specifically, starting from each entity mention $m_i \in M$, we desire the model to learn 

Starting from each mentioned entity $m_i$, we desire the model to learn to walk over the graph to retrieve relevant knowledge and form a $T$-length reasoning path for answering this question, where $T$ is a hyper-parameter denote the 
longest reasoning path required to answer the questions.
% Starting from each mentioned entity $m_i$, we desire the model to learn to walk over the graph to retrieve relevant knowledge for answering this question.
We define each reasoning path starting from the entity mention $m_i$ as a chain of entities (states) random variables $\rho_i = \{e^{t}_i\}_{t=0}^T$, where each mentioned entity is the initial state, i.e., $e^{0}_i = m_i$. The union of all paths for this question is defined as $\Rho = \{ \rho_i \}$, which contains the reasoning paths from each mentioned entity to answer the question.

% $\pib_i^{(0)} \xrightarrow{\rb_i^{(1)}} \pib_i^{(1)} \ldots  \pib_i^{(T-1)}\xrightarrow{\rb_i^{(T)}} \pib_i^{(T)}$ starting from each entity mention $m_i \in M$. At $t$-th reasoning step, each $\pib^{(t)}_i = \Pb_{ent}^{(t)}(e | q, m_i) \in \mathcal{R}^{|\mathcal{E}|}$
% is a probability distribution, with $\pib^{(t)}_i[e]$ being the probability of staying at entity $e$ 
% % is a entity distribution row-vector
% is a probability vector representing the current entity distribution starting from mention $m_i$, 
% and $\rb_i^{(t)}  \in \mathcal{R}^{|\mathcal{R}|}$ 
% % is a relation action probability row-vector. 
% % \YX{Are all the vectors in this paper row vectors? It will be good if we can be consistent. If so, just add note saying that all the vectors are row vectors.}
% is a probability vector representing the current relation distribution predicted to transit the state.

% Take the first reasoning step as an example, the state starting from entity mention $m$ = ``Bauhaus'' has highest probability staying at ``Walter Gropius'', who is the inventor of ``Bauhaus'' as its representative. And by knowing ``Walter'' was involved in the ``World War I'' in the second reasoning step, \method could infer the correct answer, via a path $\text{Bauhaus} \xrightarrow{\text{founded}} \text{Walter} \xrightarrow{\text{in war}} \text{World War I}$.

%For each reasoning path $\rho_i$, 
% \YS{Notation $\pi$ is overloaded. We didn't give a notation for a path yet. } 
\method factorizes $\Pb\big(a|  q, M \big)$ by incorporating  possible paths $\Rho$ as a latent variable, yielding:

%\YS{more precisely, we need to sum up all the possible paths. For each path $i, i_1, i_2, \ldots, i_T$, we do the factorization into (1) the probability to get the path; (2) the probability to arrive the answer given the path. For (1), decompose the path into T 1-step transitions (e.g., $T_{i_{t-1},i_{t}}$) }
% \YX{Please correct all occurrences of $\{m_i\}$ and $\{\pib_i\}$ to be with subsripts and upper scripts: $\{m_i\}_{i=1}^n$ and $\{\pib_i\}_{i=1}^n$. You can also use $M$ and $\Pbi$ directly. Also in Sec 3.2 you defined $|M|=N$; you can define it here.}
% {\small
% \begin{align}
% &\Pb\big( a |  q, M \big) = \Pb \big(\Pib | q, \{m_i\} \big) \cdot  \Pb\big(a | q, M, \Pib\}\big) \\
% &=\Big( \prod_{m_i} \Pb\big(\pib_i^{1:T} | q, m_i \big) \Big) \cdot  \Pb\big(a | q, \{m_i, \pib_i^{1:T}\}\big)    \\
% &=  \Big(\prod_{m_i} \prod_{t=1}^T \underbrace{\Pb\big(\pib_i^{t} |q, m_i, \pib_i^{<t} \big)}_\text{1) $\KG$ Reasoning} \Big)  \underbrace{  
% \Pb\big(a | q, \{m_i, \pib_i^{1:T}\}\big)}_\text{2) knowledge-injected \texttt{LM}} \nonumber \label{eq:overall}
% \end{align}
% }

{\small
% \vspace{-10pt}
\begin{align*}
&\Pb\big( a |  q, M \big) = \sum\nolimits_{\Rho}  \Pb \big(\Rho| q, \{m_i\}_{i=1}^n \big) \cdot  \Pb\big(a | q, M, \Rho\big) \\
&=\sum\nolimits_{\Rho}  \Big( \prod_{i=1}^{n} \Pb\big(\rho_i | q, m_i \big) \Big) \cdot  \Pb\Big(a | q, \{m_i, \rho_i\}_{i=1}^{n} \Big)    \\
&= \sum\nolimits_{\Rho}   \Big(\prod_{i=1}^{n} \prod_{t=1}^T \underbrace{\Pb\big(e_i^{t} |q, e_i^{<t} \big)}_\text{$\KG$ Reasoning (\ref{sec:reason})} \Big)  \underbrace{  
\Pb\Big(a | q, \{e_i^{0:T}\}_{i=1}^{n} \Big)}_\text{knowledge-injected \texttt{LM} (\ref{sec:lm})}  \label{eq:overall}
\end{align*}
}% \YX{(add explanation) here the second equation is because we assume each $\pib$ to be independent with each other and with other entities, i.e., $\Pb(\pib_i^{1:T}|q, M)=\Pb(\pib_i^{1:T}|q, m_i)$.}
%The second and third equation assume that reasoning path starting from different entities are mutually independent conditioned on question: $e_i^{(t)} \indep e_j^{(t)} \mid q, \forall t$.

We assume (1) reasoning paths starting from different entities are generated independently; and (2) reasoning paths can be generated autoregressively.

% \looseness=-1 
%\YS{the previous paragraph needs update. Need to define $\pi$ and explain why do we have these constraints, and how to implement these constraints.} \YS{I think what you try to do here is to justify the decomposition.  }

In this way, the QA problem can be decomposed into two entangled steps: 1) $\KG$ Reasoning, which autoregressively walks through the graph to get a path $\rho_i$ starting from each entity mention $m_i$; and 2) knowledge-injected \texttt{LM}, which benefits from the reasoning paths to obtain the out-context knowledge for answer prediction. 

The relational path $\rho_i$ in $\KG$ Reasoning requires the selection of next entity $e_i^t$ at each step $t$. We further decompose it into two steps: 1.a) relation prediction, in which \texttt{LM} is involved to predict the next-hop relation 
%next-hop continuous relation action 
based on the current state and context; and 1.b) the non-parametric state transition, which is to predict the next-hop entity based on the $\KG$ and the predicted relation. 
%to updating entity state. 
Formally:

% \vspace{-10pt}
{
\footnotesize
\begin{align}  \nonumber
\underbrace{\Pb\big(e_i^{t} | q, e_i^{<t}\big)}_\text{$\KG$ Reasoning (\ref{sec:reason})} \!=\! \sum_r \underbrace{\Pb_{rel}\big(r_i^{t} | q,  e_i^{<t}\big)}_\text{relation prediction (\ref{sec:relation})} \cdot \underbrace{\Pb_{walk}\big(e_i^{t} | r_i^{t}, e_i^{<t}\big)}_\text{contextualized random walk (\ref{sec:crw})}
\end{align}  
}

%\YS{it seems to me $\underbrace{\Pb\big(\pib_i^{t} | q, m_i, \pib_i^{<t}\big)}_\text{$\KG$ Reasoning (\ref{sec:reason})}$ should be  $\underbrace{\Pb^{(t)}\big(e_j | q, m_i, \pib_i^{<t}\big)}_\text{$\KG$ Reasoning (\ref{sec:reason})}$. In other words, the distribution is not defined over $\pi$ but over entities $e_j$}

%To support differentiable learning, in our \method framework, we do not really do discrete state transition. Instead, 
We keep track of the entity distribution at each step $t$ via the probability vector\footnote{Throughout the paper, all vectors are row-vectors} $\pib^{(t)}_i \in \mathcal{R}^{|\mathcal{E}|}$,
with $\pib^{(t)}_i[e]$ being the probability of staying at entity $e$, i.e., $\Pb\big(e_i^{t} = e | q, e_i^{<t}\big)$. 
%\YS{We use $\pib^{(t)}_i[e]$ to denote $\Pb\big(e_i^{t} = e | q, e_i^{<t}\big)$. Please confirm whether it is the case.}

% More specifically, 
% % maintains states for each entity mention and conduct walking. 
We highlight the three procedures in red dotted box in Figure~\ref{fig:framework}. We take the first reasoning step starting from entity mention ``Bauhaus'' as an example. In the first red box within \texttt{KIL}, we predict which relation action should be taken for entity ``Bauhaus'', and send the prediction (e.g. ``Founded'') to $\KG$. In the second red box, $\KG$ re-weights the graph and conducts contextualized random walk to update entity distribution, where ``Walter'' has the highest probability. Finally, weighted by the entity distribution, an aggregated entity embedding is sent back to \texttt{KIL} and added into a placeholder token as the knowledge, so the later \texttt{LM} layer knows to focus on the retrieved ``Walter''. We introduce these steps in the following.

% Take the first reasoning step as an example, the state starting from entity mention $m$ = ``Bauhaus'' has highest probability staying at ``Walter Gropius'', who is the inventor of ``Bauhaus'' as its representative. And by knowing ``Walter'' was involved in the ``World War I'' in the second reasoning step, \method could infer the correct answer, via a path $\text{Bauhaus} \xrightarrow{\text{founded}} \text{Walter} \xrightarrow{\text{in war}} \text{World War I}$.

% Firstly, \texttt{KIL} predicts relation action to take for each entity mention, based on which $\KG$ re-weights the graph and conduct contextualized random walk to update entity distribution state. An weighted aggregated entity embedding will then be sent back to \texttt{KIL} and add into a placeholder token as retrieved knowledge. By repeating such procedure multiple times, \method could retrieve knowledge from $\KG$ that are multi-hop away from initial entities in the question, based on which to answer complex open-domain questions.

\paragraph{Input}
% \YX{I slightly changed the wording here.} 
Initially, we first identify all $N$ entity mentions $\{m_i\}_{i=1}^N$ in the input question $q$ as well as the corresponding $\KG$ entities\footnote{
For Wikipedia pretraining, we use the ground-truth entity label as one-hot initialization for $\pib_i^0$. For downstream tasks we use GENRE~\citep{DBLP:conf/iclr/CaoI0P21} to get top 5 entity links.
% We could flexibly utilize ground-truth entity label or trained model such as GENRE~\citep{DBLP:conf/iclr/CaoI0P21} to get entity linking.
}..
% We define the initial entity distribution for $m_i$ as $\pib_i^0 = \Pb(e | m_i, q)$
For each mention $m_i$ we add three special tokens as the interface for Knowledge Interaction Layers (\texttt{KIL}) to send instruction and receive knowledge: we add a [\texttt{S-ENT}] token before, and [\texttt{REL}], [\texttt{T-ENT}] tokens after each entity mention $m_i$. \texttt{KIL} can be flexibly inserted into arbitrary $\texttt{LM}$ intermediate layer. By default, we just insert each \texttt{KIL} every $N$ 
% \YX{$N$ is also the number of entity mentions. Can you change to another letter?} 
Transformer-based $\texttt{LM}$ layers, thus the input to the $t$-th \texttt{KIL} are contextualized embeddings of each token $k$ as $\texttt{LM}^{(t)}_k$, including added special tokens.

% Research have shown that a single pre-trained Language Model (\texttt{LM}, e.g., T5) already stores a portion of knowledge within its parameter to solve some single-hop questions~\citep{DBLP:conf/emnlp/RobertsRS20}, but its performance is still lower than retrieval-based method that takes Wikipedia as knowledge source, and cannot handle multi-hop questions. 

\subsection{\texttt{LM} involved $\KG$ Reasoning} \label{sec:reason}
\noindent We first introduce the reasoning process $\Pb\big(e_i^{t} | q, e_i^{<t}\big) \!=\! \sum_r \Pb\big(r_i^{t} | q,  e_i^{<t}\big) \cdot\Pb\big(e_i^{t} | r_i^{t}, e_i^{<t}\big)$.
% \looseness=-1
% following Eq.(\ref{eq:reason}).
\subsubsection{Relation Prediction.} \label{sec:relation}
% \YX{I slightly changed the wording here.} 
\noindent For each entity mention $m_i$, we desire to predict which relation action should take $r_i^{t}$ as instruction to transit state. 
We define the predicted relation probability vector $\rb_i^{(t)} = \Pb_{rel}\big(r_i^{t} | q,  e_i^{<t}\big) \in \mathcal{R}^{|\mathcal{R}|}$ representing the relation distribution to guide walking through the graph. 
Denote the corresponding [\texttt{REL}] token as $\texttt{REL}[i]$ (and similarly for other special tokens). The contextual embedding $\texttt{LM}^{(t)}_{\texttt{REL}[i]}$ encode the relevant information in question $q$  %\YS{what is x?} 
that hints next relation. We maintain a global relation key memory $\texttt{K}_{rel} \in \mathbb{R}^{|\mathcal{R}| \times d}$ storing each relation's $d$-dimentional embedding. To calculate similarity, we first get relation query $Q^{(t)}_{\texttt{REL}[i]}$ by projecting relation token's embedding into the same space of key memory via a projection head Q-Proj\footnote{We denote a non-linear MLP projection as X-Proj$(h)=W^X_2 \sigma(W^X_1h+b_1)+b_2$, where X have different instantiations.} followed by a LayerNorm (abbreviated as LN), and then calculate dot-product similarity followed by softmax:
% \looseness=-1
\begin{align}
        &Q^{(t)}_{\texttt{REL}[i]} = \text{LN}^{(t)}\big(\text{Q-Proj}^{(t)}(\texttt{LM}^{(t)}_{\texttt{REL}[i]})\big), \\
    &\rb_i^{(t)} =\Pb_{rel}\big(r_i^{t} | q,  e_i^{<t}\big) = \text{Softmax} \big( Q^{(t)}_{\texttt{REL}[i]} \ \texttt{K}_{rel}^T   \big). 
\end{align}
%\YS{from the above formula,  $\pib_i^{<t}$ is not used to calculate the relation probability. Instead, it requires other intermediate results from language models.}

Note that the relation queries $\texttt{LM}^{(t)}_{\texttt{REL}[i]}$ are different for every mention $m_i$ and reasoning step $t$ depending on the context, and thus the the relation distributions $\rb_i^{(t)}$ gives contextualized predictions based on the question $q$. The predicted relations are sent to the knowledge graph reasoning module as instruction to conduct state transition.

% and embedding $\texttt{LM}^{(t)}_{\texttt{S-ENT}}$ should identify the current entity state, $\texttt{LM}^{(t)}_{\texttt{REL}}$ should identify the next relation ,acccmtion to take, and the retrieved knowledge (aggragated entity embedding) will be added back to $\texttt{LM}^{(t)}_{\texttt{T-ENT}}$ for passing the knowledge graph reasoning's results back to \texttt{LM}. 

% We denote each added token as a reasoning prompt. \method could simultaneously proceed Reasoning for all prompt within the context. For simplicity, we only focus on one specific entity $e$ to show the procedure.
% Before each reasoning step, we use $N$ Transformer layer of $\texttt{LM}$ to process the contexts. We denote $\texttt{LM}^{(t)}_{\texttt{S-ENT}}$, $\texttt{LM}^{(t)}_{\texttt{REL}}$, $\texttt{LM}^{(t)}_{\texttt{T-ENT}}$ as the three $d$-dimensional embedding vector before $t$-th reasoning step for the three token .

% For both pre-training and fine-tuning, \method takes some paragraphs $x$ as input and learns to predict outputs $y$ via $P(y|x)$. For pre-training, the task is to predict or autoregressively generate masked tokens, while for fine-tuning on QA, $x$ is question and $y$ is the answer.

% \begin{algorithm}
%   \caption{Pseudo Codes of Contextualized Random Walk (CRW) Implementation}
%   \Function{preprocess($\pib^0 = \{\pib_i^0 \}_{i=1}^N$, $A$)}{
%   }
% \end{algorithm}

\subsubsection{Contextualized KG Random Walk} \label{sec:crw}

% Many previous knowledge graph reasoning works propose to walk over the graph where each state is a discrete entity node. Such methods cannot be differentiated and they use reinforcement learning to get reward for training the walking agent. Our approach, instead, propose to keep a continuous entity distribution as the state, i.e., $\pib^{(t)} = \Pb^{(t)}_{ent}(e | m_i, x)$, thus the transition policy could be directly optimized through back-propagation. 
% \YX{Slightly changed the notation here; please check if it makes sense.}
\noindent Next, we introduce how we conduct state transition $\Pb_{walk}\big(e_i^{t} | r_i^{t}, e_i^{<t}\big)$. One classic transition algorithm is random walk, which is a special case of markov chain, i.e. the transition probability only depends on previous state. Consider a state at entity $s$, the probability walking to target $t$ is $\frac{1}{deg(s)}$ if $A[s,t]=1$. Based on it, we define the Markov transition matrix for random walk as $M_{rw} = \Db_A^{-1}A$, where the degree matrix $\Db_A\in \mathbb{R}^{|\mathcal{E}| \times |\mathcal{E}| }$ is defined as the diagonal matrix with the degrees $deg(1),\ldots,deg({|\mathcal{E}|)}$ on the diagonal.
With random walk Markov matrix $M_{rw}$ we can transit the state distribution as:
$\pib^{(t)} = \pib^{(t-1)} M$,
The limitation of random walk is that the transition strategy is not dependent on the question $q$. 
%\YS{x?} 
We thus propose a Contextualized Random Walk (\rw). 
% \looseness=-1

Based on the predicted relation distribution $\rb_i^{(t)}$, we calculate a different weighted adjacency matrix $\widetilde{A}_i^{(t)} \in \mathbb{R}^{|\mathcal{E}| \times |\mathcal{E}|}$ by adjusting the edge weight:
\begin{align}
    % &\widetilde{A}_i^{(t)} = \sum_{r  \in \mathcal{R}}  w_r  \cdot \rb_{i,r}^{(t)} \ \big(D_A^{-1}A_r\big)  \\
    &\widetilde{A}_i^{(t)} = \sum\nolimits_{r  \in \mathcal{R}}   w_r \cdot  \rb_{i,r}^{(t)}  \cdot  A_r,\\
    & {M_{crw, i}}^{(t)} =  \Db_{\widetilde{A}_i^{(t)}}^{-1} \widetilde{A}_i^{(t)}, \ \ \forall i \in [1, N].
\end{align}
where $w_r$ is a learnable importance weight for relation $r$ that helps solving downstream tasks, and $\rb_{i,r}^{(t)}$ is the probability corresponding to relation $r$ in $\rb_{i}^{(t)}$. With the transition matrix ${M_{crw,i}}^{(t)}$, the state transition is defined as $\pib^{(t)}_i = \pib^{(t-1)}_i  M_{crw,i}^{(t)}$.

% \YS{(1) the double normalization mentioned above seems redundant. Only the second one is needed. (2) writing wise, try to use high-level notation to help people to get a high-level idea, such as mentioning transition matrix that is specific to the starting point $i$ and the current step $t$. Then define this transition matrix. Then write down the calculation of $\pi_i^{(t)}$. (3) another confusing point is, $\pi_i^{(t)}$ is computed in a deterministic way. why do we need to use $P(\pi_i^{(t)}|...)$ notation for $\pi_i^{(t)}$? }

\rw allows each reasoning path $\rho_i$ to have its transition matrix. However, as the total number of entity nodes $|\mathcal{E}|$ could be huge (e.g., 5M for WikiData), we cannot afford to update the entire adjacency matrix for every in-batch mention. We thus adopt a scatter-gather pipeline to implement graph walking as shown in Algorithm~\ref{algo:rw}. We first gather the entity and relation probability to each edge, and then scatter the probability to target nodes. This allows us to simultaneously conduct message passing with modified adjacency weight $\widetilde{A}_i^{t}$ for all entity mention $m_i$ in parallel. 

\begin{algorithm}[!ht] 
\lstset{style=style_snippet}
\begin{lstlisting}[language=Python]
def ContextualizedRandomWalk(
    i_init, KG,   # initial entity index and Graph
    w_deg, w_rel, # inv(degree) and relation weights 
    p_ent, p_rel  # entity and predicted relation dis-                          
                  # tribution tensor @ t-th step.
): -> FloatTensor 
    # Get <src, rel, tgt> edge list of k-hop subgraph
    i_src, i_rel, i_tgt = k_hop_subgraph(i_init, KG)
    # Gather entity and relation probability to edge
    p_src  = (p_ent * w_deg)[:, i_src] # N x n_edge
    p_rel  = (p_rel * w_rel)[:, i_rel] # N x n_edge
    p_edge = l1_normalize(p_src * p_rel, dim=1)
    # Scatter edge probability to target node
    p_ent  = scatter_add(src=p_edge, idx=i_tgt, dim=1)
    return p_ent  #(t+1)-th step's entity distribution
\end{lstlisting}
\caption{Pytorch Pseudocode of CRW}\label{algo:rw}
\end{algorithm}

The complexity is $\#$ of in-batch entities times $\#$ of edges in $T$-hop subgraph starting from these entities, i.e., $\mathcal{O}(n \times \#\text{edge})$, and thus this operation is not expensive. 
% \YX{Is this local subgraph for only one example or for one batch? Please specify. If the latter, also specify the batch size here.}
Another concern is why not using Graph Neural Networks (GNNs). We provide discussion in Sec.~\ref{sec:gnn} in Appendix.

\subsection{Knowledge-Injected \texttt{LM}} \label{sec:lm}

\noindent After we get the updated entity distribution $\pib^{(t)}_i$, we want to inject such information back to the $\texttt{LM}$ without harming its overall structure. We maintain a global entity embedding value memory $\texttt{V}_{ent} \in \mathbb{R}^{|\mathcal{E}| \times d}$ storing entity embeddings. We only consider the entities within the sampled local subgraph in each batch. We thus get an entity index list $\Ib$ as the query to sparsely retrieve a set of candidate entity embeddings and then aggregate them weighted by entity distribution and embedding table. We then use a Value Projection block to map the aggregated entity embedding into the space of $\texttt{LM}$, and then directly add the transformed embedding back to the output of \texttt{T-ENT}.
\begin{align} 
    &V^{(t)}_i = \text{V-Proj}^{(t)}\big(\pib^{(t)}_i \cdot \texttt{V}_{ent}[\Ib]\big),\\
    &\widehat{\texttt{LM}}^{(t)}_{\texttt{T-ENT}[i]} =  \text{LN}^{(t)}\big(\texttt{LM}^{(t)}_{\texttt{T-ENT}[i]} + V^{(t)}_i \big).
\end{align}
Then, we just take all $\widehat{\texttt{LM}}^{(t)}_{\texttt{T-ENT}}$ as input to next Transformer-based \texttt{LM} layer to learn the interaction between the retrieved knowledge with in-context words via self-attention. 

By repeating the \texttt{KIL} for $T$ times, the final representation $\widehat{\texttt{LM}}^T$ is conditioned on the reasoning paths $\rho_i = e_i^{0:T}$, which reaches entities that are $T$-hop away from initial entity $m_i$ in the question. Finally, we can predict the answer of open questions $\Pb\big(a | q, \{e_i^{0:T}\}_{i=1}^{n} \big)$ by taking knowledge-injected representation $\widehat{\texttt{LM}}^T$ for span extraction, entity prediction or direct answer generation. 
% \YS{the previous paragraph is outdated.}

\subsection{Pre-Train \method to Reason}

%\YS{I feel there are lots of smart designs for the pre-training tasks. But the writing does not reflect that yet.}\YS{try to write: issues; solutions; losses}

\noindent The design of \method allows end-to-end training given QA datasets. However, due to the small coverage of knowledge facts for existing QA datasets, we need to pretrain \method on a large-scale corpus to get good entity embeddings.
% it's unlikely to directly train \method using downstream QA tasks without proper initialization. 

% We therefore introduce how we pre-train \method using unlabelled Wikipedia Corpus to learn to reason over $\KG$.

\paragraph{Salient Span Masking} 
% \YS{shall we upgrade this into a subsection? right now we only have one subsection.}
One straightforward approach is to use
Salient Span Masking (SSM) objective~\citep{DBLP:journals/corr/abs-2002-08909} masks out entities or noun tokens requiring specific out-of-context knowledge. We mainly mask out entities for guiding \method to reason. Instead of randomly masking entity mentions, we explicitly sample a set of entity IDs and mask every mentions linking to these entities. This could prevent the model copy the entity from the context to fill in the blank. We also follow~\citep{DBLP:conf/nips/YangDYCSL19} to mask out consecutive token spans. We then calculate the cross-entropy loss on each salient span masked (SSM) token as $\mathcal{L}_{SSM}$. 
% \looseness=-1

%\YS{why these two objectives? How these two objective connect to the two losses mentioned below.}

\subsubsection{Weakly Supervised Training of $\texttt{KIL}$}
\noindent Ideally, \method can learn all the entity knowledge and how to access the knowledge graph by solely optimizing $\mathcal{L}_{SSM}$. However, without a good initialization of entity and relation embeddings, \texttt{KIL} makes a random prediction, and the retrieved entities by $\KG$ reasoning are likely to be unrelated to the question. In this situation, \texttt{KIL} does not receive meaningful gradients to update the parameters, and \texttt{LM} learns to ignore the knowledge.  
%\YS{what are the above two paragraphs? Can we put them into subsections related to specific losses? In other words, quickly tell the readers what they are reading.}
% no matter what entities retrieved within the reasoning procedure, the final aggregated embeddings could not benefit the language model objective, and thus the model won't get the correct signal to learn which relation should choose, and how to improve entity embeddings.
% \YX{This is not quite true since you also train the entity and relation embeddings. I'd suggest to weaken the tone. Like: However, if the entity and relation embeddings are not sufficiently trained, the model might not benefit from the \texttt{KIL} layers.} 
To avoid this cold-start problem and provide entity and relation embedding a good initialization, We utilize the following two external signals as self-supervised guidance.
% \looseness=-1

% \vspace{0.05in}
\paragraph{Entity Linking Loss}
To initialize the large entity embedding tables in $\texttt{V}_{ent}$, we use other entities that are not masked as supervision. Similar to~\citet{DBLP:journals/corr/abs-2004-07202}, we force the output embedding of [\texttt{S-ENT}] token before the first \texttt{KIL} followed by a projection head E-Proj to be close to its corresponding entity embedding:
\begin{align*}
    &E_{\texttt{S-ENT}[i]} = \text{LN}\big(\text{E-Proj}(\texttt{LM}^{(1)}_{\texttt{S-ENT}[i]})\big), \\
    &\Pb^{(0)}_{ent}\big(e | m_i, q \big) = \text{Softmax} \big( E_{\texttt{S-ENT}[i]}\ \texttt{V}_{ent}[\Ib]^T   \big),\\
    &\mathcal{L}_{ent} = \sum\nolimits_{m_i} -\log \Pb^{(0)}_{ent}\big(e | m_i, q \big) \cdot \pib^0_i[\Ib].
\end{align*}
% \YX{Is this softmax here over all entities? It is in-batch entities if I remember correctly. Either case, please specify it.}
Similar to Section~\ref{sec:lm}, we only consider entities within the batch, denoted by index $\Ib$.
This contrastive loss guides each entity's embedding $\texttt{V}_{ent}[e]$ closer to all its previously mentioned contextualized embedding, and thus memorizes those context as a good initialization for later knowledge integration. 
% \looseness=-1

% \vspace{0.05in}
\paragraph{Weakly Supervised Relation Path Loss}
Entity mentions within each Wikipedia passage are naturally grounded to WikiData $\KG$. Therefore, after we mask out several entities, we can utilize the $\KG$ to get all reasoning paths from other in-context entities to the masked entities as weakly supervised relation labels.
% \looseness=-1

\begin{figure}[t!]
    \centering
    \includegraphics[width=1.0\columnwidth]{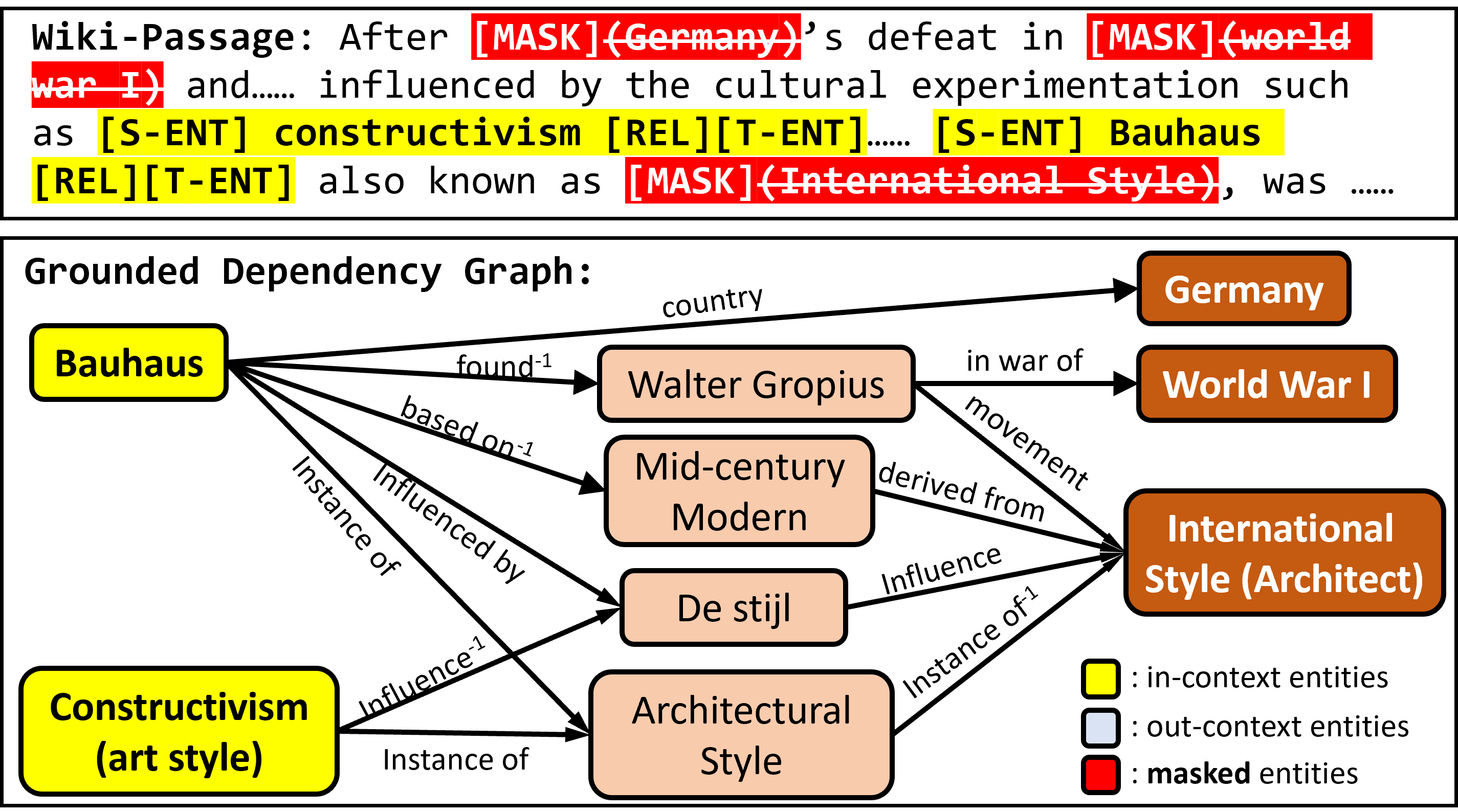}
    \caption{Pre-training sample w/ golden reasoning path. More real examples are shown in Table~\ref{tab:pretrain1} in Appendix.}
    \label{fig:pretrain_example}
%  \vspace{-.1in}
\end{figure}

Formally, we define a \textbf{Grounded Dependency Graph} $\mathcal{DG}$, which contains all reasoning paths within $T$-step from other in-context entities to masked entities, and then define $R_{\mathcal{DG}}(m_i, t)$ as the set of all relations over every edges for entity mention $m_i$ at $t$-th hop. Based on it, we define the weakly supervised relation label $q^{(t)}_i \in \mathbb{R}^{|\mathcal{R}|}$ as the probabilistic vector which uniformly distributed on each relation in set.
% , i.e., $q^{(t)}_i[r] = \frac{1}{|R_{\mathcal{DG}}(m_i, t)|}, \text{if }\ r \in R_{\mathcal{DG}}(m_i, t)$.
% $q^{(t)}_i[r] = \begin{cases}
%     \frac{1}{|R_{\mathcal{DG}}(m_i, t)|},& \text{if }\ r \in R_{\mathcal{DG}}(m_i, t)\\
%     0,              & \text{otherwise}
% \end{cases}$
Note that we call uniformly-weighted $q^{(t)}_i$ as weakly supervised because 1) some paths 
lead to multiple entities rather than only the target masked entity; 2) the correct relation is dependent on the context. Therefore, $q^{(t)}_i$ only provides all potential candidates for reachability, and more fine-grained signals for reasoning should be learned from unsupervised $\mathcal{L}_{SSM}$. We adopt a list-wise ranking loss to guide the model to assign a higher score on these relations than others.
\begin{align*}
\mathcal{L}_{rel} = \sum\nolimits_{m_i}\sum\nolimits_{t=1}^{T} -\log \Pb^{(t)}_{rel}\big(r | m_i, q \big) \cdot q^{(t)}_i.
\end{align*}

% As an example in Figure~\ref{fig:pretrain_example}
% , we take one paragraph from ''Bauhaus'' and mask out entities ''Germany`` and ''World War I`` as targets. We then construct the grounded dependency graph from other entities to these two masked entities and get the weakly supervised relation label for each in-context source entity per reasoning step. More real examples are shown in Table~\ref{tab:pretrain1} in Appendix.

Overall, $\mathcal{L}_{ent}$ and $\mathcal{L}_{rel}$  provide \method with good initialization of the large $\KG$ memory. Afterward, via optimizing $\mathcal{L}_{SSM}$, the reasoning paths that provide informative knowledge receive a positive gradient, guiding \method to reason.

% Specifically, for each Wikipedia passage containing $N$ entity mentions, we first construct a \textbf{K-hop Induced Subgraph}, in which all the paths with length shorter than $K$ connecting any pair of entities are retained. Then, during training, after we randomly mask out a set of entities, we could easily calculate a 

% spanning tree starting from the masked entities to all other in-context entities. Then, for each entity mention $m_i$, 

% To pre-train the model to reason, we desire a self-supervised task that the model could not simply answer by looking at local contexts, but require external knowledge. To serve the purpose, besides the vanilla masked language model, we specifically sample a portion of entities within each page, and mask out all their mentions. In this way, the model cannot simply copy the tokens from contexts to fill in the blank. In addition, as we have the grounded entity graph, we could pre-extract all paths from in-context entity mentions to target masked entities. These paths could serve as golden ground-truth for relation prediction during pre-training, guiding our model to make reasonable relation prediction at an early stage. As illustrated in Figure~\ref{fig:pretrain_example}, we mask out two entities Mudhoney and Nirvana from the paragraph, which is hard for a vanilla \texttt{LM} to generate based on context. We also provide the one-hop and two-hop paths from in-context entity mentions.

%% file: sections/experiments.tex
\begin{table}[t]
\centering
\scriptsize
\begin{tabular}{l|ccc} \toprule
\textbf{Name}   & \textbf{Number}  & \textbf{dimension} & \textbf{\#param (M)} \\ \midrule
Number of  Entity& 4,947,397 & 128 & 633\\ 
Number of  Relation& 2,008  & 768 & 1.5\\ 
Number of  Edges & 45,217,947 & - & 47\\ 
\bottomrule
\end{tabular}
\caption{Statistics and parameter of $\mathcal{KG}$ Memory.}
\label{tab:stat}
\end{table}

\begin{table*}[ht]
\hspace{-0.2in}
\centering
\small
\begin{tabular}{lc|ccc|cc} \toprule
\textbf{Models}   & \textbf{\#param}  & \textbf{NQ} & \textbf{WQ} & \textbf{TQA}  & \textbf{ComplexWQ}  & \textbf{HotpotQA}  \\ \midrule
T5 (Base)           & 0.22B  &  25.9 & 27.9 & 29.1  &  11.6 & 22.8\\ 
 \textbf{+ \method} ($T$=1) & 0.23B + \underline{0.68}B &  28.3 & 30.6 & 32.4  &  20.8  & 24.1 \\ 
 \textbf{+ \method} ($T$=2) & 0.24B + \underline{0.68}B  &  28.9 & 31.2 & 33.7  &  23.7 & 26.3\\ \midrule
T5 (Large)           & 0.74B  &  28.5 & 30.6 & 35.9  &   16.7 & 25.3\\ 
 \textbf{+ \method} ($T$=1) & 0.75B + \underline{0.68}B   &  30.6 & 32.8 & 39.1 & 24.5 & 28.2 \\ 
 \textbf{+ \method} ($T$=2) & 0.76B + \underline{0.68}B   &  \textbf{31.0} & \textbf{34.3} & \textbf{40.0}  & \textbf{27.1}  & \textbf{31.4}\\ \midrule
T5-3B~\citep{DBLP:conf/emnlp/RobertsRS20}              & 3B    &  30.4 & 33.6 & 43.4  &   - & 27.8\\ 
T5-11B~\citep{DBLP:conf/emnlp/RobertsRS20}              & 11B    &  32.6 & 37.2 & 50.1 &  - & 30.2 \\ 
\bottomrule
\end{tabular}
\caption{\textbf{Closed-Book Generative QA} performance of Encoder-Decoder \texttt{LM} on Single- and Multi-hop Dataset. }
\label{tab:close}
\end{table*}

% In this section, we evaluate our method for \textit{Closed-Book} settings

% Most existing ODQA systems follow two pipelines: 1) a closed-book setup, which assume the model's parameter stores the required knowledge, with which the model could directly conduct reasoning and generate the answer; 2) an open-book setup, which resort to retrieving passages in a large text corpus that are likely to contain missing knowledge, and then use a separate Language Model to re-encode question and retrieved passage and predict answer.

%\subsection{Implementation Details}

\noindent The proposed \texttt{KIL} layers can be pugged into most Transformer-based Language Models without hurting its original structure. In this paper, we experiment with both encoder-based \texttt{LM}, i.e. RoBERTa-base ($d=768, l=12$), and encoder-decoder \texttt{LM}, i.e. T5-base ($d=768, l=12$) and T5-large ($d=1024, l=24$). For all \texttt{LM}s, add 1 \texttt{KIL} layer or 2 \texttt{KIL} layers to the encoder layers. 
The statistics of $\mathcal{KG}$ are shown in Table~\ref{tab:stat}. Altogether, it takes about 0.67B parameter for $\mathcal{KG}$ memory, which is affordable to load as model parameter. We pre-train all \texttt{LM}s using the combination of $\mathcal{L}_{SSM}$, $\mathcal{L}_{ent}$ and $\mathcal{L}_{rel}$
for 200k steps on 8 V100 GPUs, with a batch size of 128 and default optimizer and learning rate in the original paper, taking approximately one week to finish pre-training of T5-large model, and 1-2 days for base model. Implementation details are elaborated in Appendix~\ref{sec:implement}.

% \YX{What is $N$ (number of transformer layers between KIL)? It is an important parameter}

% \paragraph{Fine-Tuning} In closed book setting, for encoder-based \texttt{LM} (i.e. RoBERTa), we follow the same procedure in~\cite{DBLP:conf/naacl/VergaSSC21}. Specifically, we added a [\texttt{MASK}] token after the question, and use its output embedding to predict entity ID in WikiData KB. This restricts the model so that it could only handle wiki-answerable questions. For encoder-decoder \texttt{LM} (i.e. T5), we could directly fine-tune to generate the answer. We mainly adopt the hyperparameters and setting in~\cite{DBLP:conf/emnlp/RobertsRS20}.

% 

\subsection{Evaluate for \textit{Closed-Book} QA}
\noindent \method is designed for improving \textit{Closed-Book} QA, so we first evaluate it in this setting. 
\paragraph{Generative QA Task}
Following the hyperparameters and setting in~\cite{DBLP:conf/emnlp/RobertsRS20}, we directly fine-tune the T5-base and T5-large augmented by our \method on the three single-hop ODQA datasets: Natural Question \textbf{(NQ)}~\citep{DBLP:journals/tacl/KwiatkowskiPRCP19}, WebQuestions \textbf{(WQ)}~\citep{DBLP:conf/emnlp/BerantCFL13} and TriviaQA \textbf{(TQA)}~\citep{DBLP:conf/acl/JoshiCWZ17}. 
To test \method's ability to solve complex questions, we also evaluate on two multi-hop QA datasets, i.e. \textbf{Complex WQ}~\citep{DBLP:conf/naacl/TalmorB18} and \textbf{HotpotQA}~\citep{DBLP:conf/emnlp/Yang0ZBCSM18}. Detailed dataset statistics and experimental setups are in Appendix~\ref{sec:exp_detail}.

Experimental results are shown in Table~\ref{tab:close}. We use Exact Match accuracy as the metric for all the datasets. On the three single-hop ODQA datasets, \method with 2 \texttt{KIL} blocks achieves 3.3 absolute accuracy improvement to T5-base, and 3.4 improvement to T5-large. Compared with T5 model with more model parameters (e.g., T5-3B and T5-11B), our T5-large augmented by \method could outperform T5-3B on NQ and WQ datasets. In addition, \method could use the generated reasoning path to interpret the model's prediction. We show examples in Table~\ref{tab:explain1} in Appendix.

For the two multi-hop QA datasets, the performance improvement brought by \method is more significant, i.e., 7.8 to T5-base and 8.2 to T5-large. Notably, by comparing the T5-3B and T5-11B's performance on HotpotQA (we take results from~\citep{DBLP:journals/corr/abs-2204-04581}), T5-large augmented by \method achieves 1.2 higher than T5-11B. This shows that \method is indeed very effective for improving \textit{Closed-Book} QA performance, especially for complex questions.

\paragraph{Entity Prediction Task}
Encoder-based \texttt{LM} (i.e. RoBERTa) in most cases cannot be directly used for \textit{Closed-Book} QA, but more serve as reader to extract answer span. 
However, \citet{DBLP:conf/naacl/VergaSSC21} propose a special evaluation setting as \textit{Closed-Book Entity Prediction}. They add a single [\texttt{MASK}] token after the question, and use its output embedding to classify WikiData entity ID. This restricts that answers must be entities that are covered by WikiData, which they call \textit{WikiData-Answerable} questions. We follow \citet{DBLP:conf/naacl/VergaSSC21} to use such reduced version of WebQuestionsSP \textbf{(WQ-SP)}~\citep{DBLP:conf/acl/YihCHG15} and TriviaQA \textbf{(TQA)} as evaluation dataset, and finetune the RoBERTa (base) model augmented by \method to classify entity ID. We mainly compare \method with EaE~\citep{DBLP:journals/corr/abs-2004-07202} and FILM~\citep{DBLP:conf/naacl/VergaSSC21}, which are two $\mathcal{KG}$ memory augmented \texttt{LM}. We also run experiments on KEPLER~\citep{DBLP:journals/corr/abs-1911-06136}, a RoBERTa model pre-trained with knowledge augmented task.
% \looseness=-1

Experimental results are shown in Table~\ref{tab:roberta}. Similar to the observation reported by \citet{DBLP:conf/naacl/VergaSSC21}, adding $\mathcal{KG}$ memory for this entity prediction task could significantly improve over vanilla $\texttt{LM}$, as most of the factual knowledge required to predict entities are stored in $\mathcal{KG}$. By comparing with FILM~\citep{DBLP:conf/naacl/VergaSSC21}, which is the state-of-the-art model in this setup, \method with reasoning step ($T=2$) outperforms FILM by 2.9, with smaller memory consumption.

\subsection{Analyze $\mathcal{KG}$ Reasoning Module}
\noindent In our previous studies, we find that using a higher reasoning step, i.e. $T=2$, generally performs better than $T=1$. We hypothesize that the $\mathcal{KG}$ we use has many missing one-hop facts, and high-order reasoning helps recover them and empowers the model to answer related questions.
To test whether \method indeed can infer missing facts, we use \textbf{EntityQuestions (EQ)}~\citep{DBLP:conf/emnlp/SciavolinoZLC21}, which is a synthetic dataset by mapping each WikiData triplet to natural questions. We take RoBERTa-base model augmented by \method trained on NQ as entity predictor and directly test its transfer performance on EQ dataset without further fine-tuning. 

% As the relation edge is within $\mathcal{KG}$, it should not be a hard task as long as \method could predict the correct relation label.

\begin{figure*}[ht!]
\hspace{-0.1in}
    \centering \includegraphics[width=1.02\columnwidth]{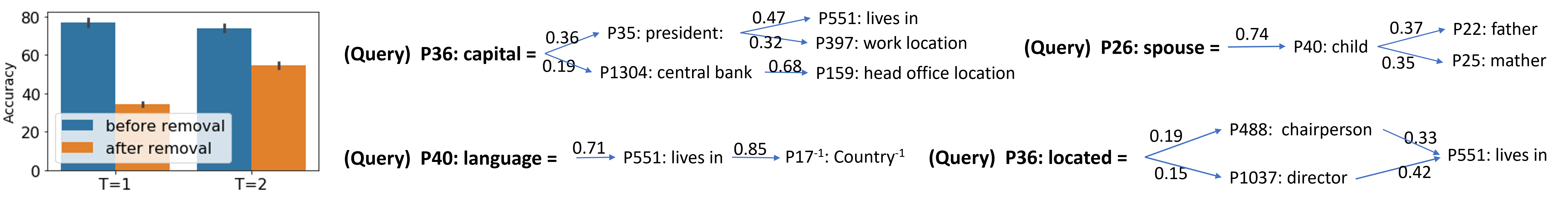}
    % \vspace{-10pt}
        \caption{\textbf{Testing the reasoning capacity of \method to infer missing relations}. On the \textbf{left}, the barplot shows the transfer performance on EQ before and after removing relation edges, \method ($T=2$) is less influenced. On the \textbf{right} shows reasoning paths (rules) automatically generated by \method for each missing relation.}
    \label{fig:analysis}
\end{figure*}

To test whether \method could recover missing relation,
we mask \textbf{all} the edges corresponding to each relation separately and make the prediction again. The average results before and after removing edges are shown on the left part of Figure~\ref{fig:analysis}. 
% Before removing the edges, both $T=1$ and $T=2$ achieve very high accuracy even without fine-tuning, as they just need to identify the correct edge from the $\mathcal{KG}$. 
When we remove all the edges to each relation, \method with $T=1$ drops significantly, while $T=2$ could still have good accuracy. To understand why \method ($T=2$) is less influenced, in the right part of Figure~\ref{fig:analysis}, we generate a reasoning path for each relation by averaging the predicted probability score at each reasoning step and pick the relation with the top score. For example, to predict the ``Capital'' of a country, the model learns to find the living place of the president, or the location of a country's central bank. Both are very reasonable guesses. Many previous works~\citep{DBLP:conf/emnlp/XiongHW17} could also learn such rules in an ad-hoc manner and require costly searching or reinforcement learning. In contrast, \method could learn such reasoning capacity for all relations end-to-end during pre-training.

\paragraph{Ablation Studies} We conduct several ablation studies to evaluate which model design indeed contributes to the model. As shown in the bottom blocks in Table~\ref{tab:roberta}, we first remove the $\mathcal{KG}$ reasoning component and provide RoBERTa base model via concatenated KB triplets and train such a model using $\mathcal{L}_{SSM}$ over the same WikiDataset. Such a model's results are close to the KEPLER results but much lower than other models with explicit knowledge memory. We further investigate the role of pre-training tasks. Without pre-training, the \method only performs slightly better than RoBERTa baseline, due to the cold-start problem of entity and relation embedding. We further show that removing $\mathcal{L}_{ent}$ and $\mathcal{L}_{ent}$ could significantly influence final performance. The current combination is the best choice to train \method to reason.

\subsection{Evaluate for \textit{Open-Book} QA}
\noindent Though \method is designed for  \textit{Closed-Book} QA, the learned model can serve as backbone for \textit{Open-Book} QA. We take DPR and FiD models as baseline. For DPR retriever, we replace the question encoder to RoBERTa + \method, fixing the passage embedding and only finetune on each downstream QA dataset. For FiD model, we replace the T5 + \method. We also changed the retriever with our tuned DPR. Results in Table~\ref{tab:open} show that by augmenting both retriever and generator, \method improves a strong baseline like FiD, for about 3.1\% for Base and 1.8\% for Large, and it outperforms the very recent KG-FiD model for 1.6\% in base setting, and achieve comparative performance in a large setting. Note that though our results is still lower than some recent models (e.g., EMDR$^2$), these methods are dedicated architecture or training framework for \textit{Open-Book} QA. We may integrate \method with these models to further improve their performance. 

\begin{table}[t]
\centering
\scriptsize
\begin{tabular}{lc|cc} \toprule
\textbf{Models}   & \textbf{\#param (B)}  & \textbf{WQ-SP} & \textbf{TQA}   \\ \midrule
 EaE~\citep{DBLP:journals/corr/abs-2004-07202} & 0.11 + \underline{0.26} & 62.4 & 24.4  \\
 FILM~\citep{DBLP:conf/naacl/VergaSSC21} & 0.11 + \underline{0.72} & 78.1 & 37.3 \\ 
KEPLER~\citep{DBLP:journals/corr/abs-1911-06136} & 0.12 &  48.3 & 24.1   \\ 
\midrule
RoBERTa (Base) & 0.12 &  43.5 & 21.3   \\ 
% RoBERTa + 2-hop KB & 120M &  43.8 & 21.0 & 15.9  \\ 
  \textbf{+ \method} ($T$=1) & 0.12 + \underline{0.68}  &  80.1 & 39.7   \\ 
  \textbf{+ \method} ($T$=2) & 0.13 + \underline{0.68} &  \textbf{80.9} & \textbf{40.3} \\ \bottomrule
\multicolumn{4}{c}{\textbf{Ablation Studies}}\\
\toprule
    RoBERTa  + Concat KB  + $\mathcal{L}_{SSM}$ & 0.12 &  47.1 & 22.6   \\ \midrule
  \textbf{+ \method} ($T$=2) \textbf{w/o PT} & 0.13 + \underline{0.68} &  46.9 & 22.7 \\  
   \ \ \ \ w. $\mathcal{L}_{SSM}$ & 0.13 + \underline{0.68} &  51.9 & 26.8 \\  
   \ \ \ \ w. $\mathcal{L}_{SSM}$ + $\mathcal{L}_{ent}$  & 0.13 + \underline{0.68} & 68.4 & 35.7 \\
    \bottomrule
\end{tabular}
% }}
\caption{\textbf{Closed-Book Entity Prediction} performance of Encoder \texttt{LM} on WikiData-Answerable Dataset. }
\label{tab:roberta}
\end{table}

\begin{table}[t]
\centering
\scriptsize
% \scriptsize
% \hspace{-.2in}
% \setlength{\tabcolsep}{0.7mm}{\scalebox{0.72}{
\begin{tabular}{lc|cccc} \toprule
\textbf{Models}    & \textbf{\#param (B)}  & \textbf{NQ} & \textbf{TQA} \\ \midrule
Graph-Retriever~\cite{DBLP:journals/corr/abs-1911-03868}    & 0.11 & 34.7 & 55.8 \\
% Path-Retriever~\cite{DBLP:conf/iclr/AsaiHHSX20}     & 0.44 & 31.7 &  -   \\
REALM~\cite{DBLP:journals/corr/abs-2002-08909}              & 0.33 + \underline{16} & 40.4 &  -   \\ \midrule
DPR~\cite{DBLP:journals/corr/abs-2004-04906} + BERT & 0.56 + \underline{16} & 41.5 & 56.8 \\ 
 \textbf{+ \method} (DPR,\ $T$=2) & 0.57 + \underline{17}  &  43.7 & 58.5 \\ \midrule
% & RAG~\citep{DBLP:conf/nips/LewisPPPKGKLYR020}                & 626M & 44.5 & 56.1 \\ 
% & Joint Top-K~\citep{DBLP:conf/nips/SachanRHDY21}       & 440M & 49.2 & 64.8 \\ 
FiD (Base) = DPR + T5 (Base)           & 0.44 + \underline{16} & 48.2 & 65.0 \\
 \textbf{+ \method} (T5,\ $T$=2) & 0.45 + \underline{17}  &  49.3 & 67.1 \\ 
 \textbf{+ \method} (DPR \& T5,\ $T$=2) & 0.46 + \underline{17}  &  51.1 & 68.4 \\ \midrule
FiD (Large) = DPR + T5 (Large)          & 0.99 + \underline{16} & 51.4 & 67.6 \\
 \textbf{+ \method} (T5,\ $T$=2) & 0.99 + \underline{17}  &  52.4 & 68.9 \\ 
  \textbf{+ \method} (DPR \& T5,\ $T$=2) & 1.00 + \underline{17}  &  \textbf{53.2} & 69.5 \\ \midrule
 KG-FiD (Base)~\citep{DBLP:conf/acl/Yu0F0WXRY022}      & 0.44 + \underline{16}  & 49.6 & 66.7 \\ 
 KG-FiD (Large)~\citep{DBLP:conf/acl/Yu0F0WXRY022}      & 0.99 + \underline{16}  & \textbf{53.2} & 69.8 \\
 EMDR$^2$~\citep{DBLP:conf/nips/SachanRHDY21}      & 0.44 + \underline{16}  & 52.5 & \textbf{71.4} \\ \bottomrule
\end{tabular}
% }}
\caption{\textbf{Open-Book QA} Evaluation.}
\label{tab:open}
\end{table}

% BART-Large & 0.39B & 36.7 & 30.6 & - & -\\
% T5-11B & 11B & 61.0 & - & - & -\\

%% file: sections/relatedwork.tex
\noindent \textbf{Open-Domain Question Answering (ODQA)} gives QA model a single question without any context and asks the model to infer out-of-context knowledge. Following the pioneering work by~\citet{DBLP:conf/acl/ChenFWB17}, most ODQA systems assume the model can access an external text corpus (e.g. Wikipedia).
Due to the large scale of web corpus (20GB for Wikipedia), it could not be simply encoded in the QA model parameters, and thus most works propose a \textit{Retrieval-Reader} pipeline, by firstly index the whole corpus and use a \textit{retriever} model to identify which passage is relevant to the question; then the retrieved text passage concatenate with question is re-encoded by a seperate \textit{reader} model (e.g., \texttt{LM}) to predict answer. As the knowledge is outside of model parameter, \citet{DBLP:conf/emnlp/RobertsRS20} defines these methods as \textit{Open-book}, with an analogy to referring textbooks during exam. \textit{Closed-book} QA models (mostly a single \texttt{LM}) try to answer open questions without accessing external knowledge. This setting is much harder as it requires \texttt{LM} to memorize all pertinent knowledge in its parameters, and even recent \texttt{LM}s with much larger model parameters is still not competitive to state-of-the-art \textit{Open-book} models.

% \paragraph{Knowledge-Base Question Answering}
% Traditional parsing-based methods parse the question into some intermediate query (e.g., SQL language, query graphs), which can execute on a knowledge base to get answer \citep{DBLP:conf/emnlp/BerantCFL13,DBLP:conf/acl/YihCHG15,DBLP:journals/tacl/ReddyTCKDSL16,DBLP:journals/corr/abs-1709-00103,DBLP:conf/acl/LiangBLFL17}. However, existing knowledge bases suffer from low coverage of entities and relations required for open-ended questions. As an alternative, several works try to incorporate the structured knowledge into neural QA models for differentiable reasoning. \cite{DBLP:conf/emnlp/LinCCR19} and \cite{DBLP:conf/emnlp/FengCLWYR20} parse the question into a sub-graph of knowledge base, and apply graph neural networks as reasoner to extract answers. \cite{DBLP:conf/iclr/ChenLYZSL20} integrates general symbolic operations as basic units, and parse questions into compositional programs to answer general questions.

\noindent\textbf{Knowledge-augmented Language Models} 
explicitly incorporate external knowledge (e.g. knowledge graph) into \texttt{LM}~\citep{DBLP:journals/corr/abs-2010-04389}.
Overall, these approaches can be grouped into two categories:
The first one is to explicitly inject knowledge representation into language model pre-training, where the representations are pre-computed from external sources~\citep{DBLP:conf/acl/ZhangHLJSL19,DBLP:conf/aaai/LiuW0PY21,DBLP:conf/emnlp/HuSC21}.
For example, ERNIE~\cite{DBLP:conf/acl/ZhangHLJSL19} encodes the pre-trained TransE~\cite{DBLP:conf/nips/BordesUGWY13} embeddings as input.
The second one is to implicitly model knowledge information into language model by performing knowledge-related tasks, such as entity category prediction~\citep{DBLP:journals/corr/abs-2010-00796} and graph-text alignment~\cite{DBLP:conf/acl/KeJRCWSZH21}.
For example, JAKET~\citep{DBLP:journals/corr/abs-2010-00796} jointly pre-trained both the KG representation and language representation by adding entity category and relation type prediction self-supervised tasks.

There also exists several QA works using $\KG$ to help ODQA. For example, \citet{DBLP:conf/iclr/AsaiHHSX20} and \citet{DBLP:journals/corr/abs-1911-03868} expand the entity graph following wikipedia hyperlinks or triplets in knowledge base. \citet{DBLP:conf/acl/DingZCYT19} extract entities from current context via entity-linking and turn them into a cognitive graph, and a graph neural network is applied on top of it to extract answer. \citet{DBLP:conf/iclr/DhingraZBNSC20} and \citet{DBLP:journals/corr/abs-2010-14439} construct an entity-mention bipartite graph and then model the QA reasoning as graph traversal by filtering only the contexts that are relevant to the question. \citet{DBLP:conf/emnlp/LinCCR19}, \citet{DBLP:conf/emnlp/FengCLWYR20} and \citet{DBLP:conf/naacl/YasunagaRBLL21} parse the question into a sub-graph of knowledge base, and apply graph neural networks as reasoner for extracting one of the entities as the answer.

To encode knowledge (significantly smaller than the web corpus) as \emph{memory} into \texttt{LM} parameter, a line of works try compressed knowledge including QA pairs~\citep{DBLP:journals/corr/abs-2204-04581, DBLP:journals/corr/abs-2102-07033,DBLP:journals/corr/abs-2209-10063}, entity embedding~\citep{DBLP:journals/corr/abs-2004-07202} and reasoning cases~\citep{DBLP:conf/emnlp/DasZTGPLTPM21, DBLP:journals/corr/abs-2202-10610}.
There's also several works utilizing Knowledge Graph ($\KG$) to augment \texttt{LM}. FILM~\citep{DBLP:conf/naacl/VergaSSC21} turns $\KG$ triplets into memory. Given a question, \texttt{LM} retrieves most relevant triplet as answer. GreaseLM~\citep{DBLP:journals/corr/abs-2201-08860} propose to interact \texttt{LM} with $\KG$ via a interaction node.

% We discuss other related works in Sec.~\ref{sec:related} in Appendix.

% Similar to query answering on graph, answering open-domain questions also require to infer out-of-context knowledge. 
% For example, given only a question $q$ as ``The Bauhaus represented Germany's recovery from which event$?$'', the QA model is asked to predict answer $a$ ''World War I``. To correctly answer such a question, the model needs to gain knowledge about all in-context entity mentions $M=\{m_i\}$, e.g., ''Bauhaus`` and ''Germany``. 
% Such an open-domain question could be abstracted as a query $(M, q, ?)$, or a probabilistic manner $P(a | q, M)$. 

%% file: sections/conclusion.tex
% In this paper, we propose a simple yet effective pre-training framework \method. We leverage both the Wikipedia hyperlinks and Wikidata relation triplets to construct $\KG$, based on which we generate relational QA dataset. 
% We then pre-train a QA model to infer the latent relation from the question, and then conduct extractive QA to get the target answer entity. \method could improve the performance of the state-of-the-art QA frameworks, especially for questions with long-tail relations.

We presented \method, a novel model that incorporates symbolic $\mathcal{KG}$ reasoning with existing \texttt{LM}s. We showed that \method can
bring significant performance gain to open-domain QA benchmarks, both for closed-book and open-book settings, as well as encoder-only and encoder-decoder models. Additionally, \method produces reasoning paths that helps interpret the model prediction. In future, we'd like to improve \method by training to conduct more reasoning steps, supporting locial reasoning, and apply \method to a broader range of  knowledge-intensive NLP tasks.

\paragraph{Acknowledgement}
We sincerely thank anonymous reviewers for their constructive comments  to improve this paper. The project was partially supported in part by CISCO, NSF III-1705169, NSF 1937599, NASA, Okawa Foundation Grant, Amazon Research Awards, Cisco research grant,
and Picsart gift.
Ziniu is supported by the Amazon Fellowship and Baidu PhD Fellowship.

%% file: sections/limitation.tex
\paragraph{Limited Reasoning Steps}
In our experiments, we show that using reasoning step $T=2$ has better performance to $T=1$ on one-hop and multi-hop (mostly two) QA datasets. Thus, it's a natural question about whether we could extending reasoning steps more? As previous KG reasoning mostly could support very long path (with LSTM design)

Though we didn't spend much time exploring before the paper submission, we indeed try using $T=3$, but currently it didn't get better results. We hypothesize the following reasons: 1) A large portion of our current model's improvement relies on the weakly supervised relation pre-training. To do it, we construct a K-hop (K=2 now) subgraph, and sample dependency graph based on it. The larger $K$ we choose, the more noise is included into the generated relation label, in an exponential increasing speed. Thus, it's harder to get accurate reasoning path ground-truth for high-order $T$. Another potential reason is that within Transformer model, the representation space in lower and upper layer might be very different, say, encode more syntax and surface knowledge at lower layers, while more semantic knowledge at upper layers. Currently we adopt a MLP projection head, wishing to map integrated knowledge into the same space, but it might have many flaws and need further improvement.

\paragraph{Large Entity Embedding Table requires Pre-Training and GPU resources}

Our current design has a huge entity embedding table, which should be learned through additional supervision and could not directly fine-tune to downstream tasks. This is restricts our approach's usage.

\paragraph{Require Entity Linking}

Current model design requires an additional step of entity linking for incoming questions, and then add special tokens as interface. A truly end-to-end model should identify which elements to start conducting reasoning by its own without relying on external models.

\paragraph{Only support relational path-based reasoning}

Though there are lots of potential reasoning tasks, such as logical reasoning, commonsense reasoning, physical reasoning, temporal reasoning, etc. Our current model design mainly focus on path-based relational reasoning, and it should not work for other reasoning tasks at current stage.

\paragraph{Unreasonable Assumption of Path In-dependency}

When we derive equation~\ref{eq:overall}, we have the assumption that reasoning paths starting from different entities should be independent. This is not always correct, especially for questions that require logical reasoning, say, have conjunction or disjunction operation over each entity state. And thus our current methods might not work for those complex QA with logical dependencies.

%% file: sections/appendix.tex
\begin{table*}[t]
\centering
\begin{tabular}{l|ccc} \toprule
\textbf{Dataset}   & \textbf{Train}  & \textbf{Dev} & \textbf{Test} \\ \midrule
Natural Questions & 58880 & 8757, 3610\\
Trivia QA& 60413 & 8837 & 11313\\
Web Questions& 2474 & 361 & 2032\\
Complex WebQ & 27623 & 3518 & 3531\\
WebQ-SP (Wiki-answerable) & 1388 & 153 & 841 \\
FreebaseQA (Wiki-answerable) & 12535 & 2464 & 2440 \\
\bottomrule
\end{tabular}
\caption{Dataset Train/Valid/Test splits.}
\label{tab:data}
\end{table*}

\begin{table}[t]
\centering
\setlength{\tabcolsep}{1mm}{\scalebox{0.75}{
\begin{tabular}{lc|cc} \toprule
\textbf{Models}   & \textbf{\#param (B)} & WQ-SP & TQA \\ \midrule
RoBERTa (Base) & 0.12 &  47.5 & 40.3   \\ 
  \textbf{+ \method} ($T$=1) & 0.12 + \underline{0.68}  &  89.7 & 61.4   \\ 
  \textbf{+ \method} ($T$=2) & 0.13 + \underline{0.68} &  \textbf{92.4} & \textbf{66.8} \\ \bottomrule
\end{tabular}}}
\caption{\textbf{Closed-Book Entity Prediction} validation performance of Encoder RoBERTa on WikiData-Answerable Dataset. }
\end{table}

\begin{table*}[ht]
\centering
\small
% {\scalebox{}{
\begin{tabular}{lc|ccc|cc} \toprule
\textbf{Models}   & \textbf{\#param}  & \textbf{NQ} & \textbf{WQ} & \textbf{TQA}  & \textbf{ComplexWQ}  & \textbf{HotpotQA}  \\ \midrule
T5 (Large)           & 0.74B  & - & - & -  &   - & -\\ 
 \textbf{+ \method} ($T$=2) & 0.76B + \underline{0.68}B   &  33.6 & 38.9 & 42.7  & 29.6  & 35.5\\ \midrule
\end{tabular}
% }}
\caption{\textbf{Closed-Book Generative QA} validation performance of T5. }
\label{tab:close}
\end{table*}

\section{Implementation Details}\label{sec:implement}

\paragraph{Entity Linking durine pre-training}

We use the 2021 Jan. English dump of Wikidata and Wikipedia. 
For each wikipedia page, we link all entity mentions with hyperlinks to WikiData entity entry, augment all other mentions with same aliases, tokenize via each \texttt{LM}'s tokenizer and split into chunks with maximum token length allowed. We then construct induced k-hop subgraphs connecting entities within each chunk for quickly get grounded computational graph.

For entities, Wikipedia provides hyperlinks with ground-truth entity ID, but it doesn't cover all the entity mentions, mostly hyperlinks only appear when this entity appears for the first time. Therefore, we first collect all entities appeared in hyperlinks as well as their aliases stored in WikiData, and then search any mentions that have any of these alias and link it to the corresponding entity.

\paragraph{Hyperparameters}
In this work, we don't have too much hyperparmaters to be tuned, as most parameters as well as optimizing setting of LM is fixed. Our random walk part is non-parametric. The only tunable hyperparamter is hidden dimension size. We simply choose one setting, which is 128 for entity embedding, and 768 for relation embedding. The former is because entity is super large (over 5M), so we use a reletively smaller dimension size. Detailed statistics about wikidata memory is in Table~\ref{tab:stat}.

\begin{table*}[h]
\scriptsize
\begin{tabular}[t]{c|c|c|c|c} \toprule
\textbf{Title}   & \textbf{Masked Text}  & \textbf{Ground Truth} & \textbf{Dependency Graph} & \textbf{2-Hop Graph}  \\ \midrule
Poolbeg & \multicolumn{1}{p{3cm}|}{the lighthouse was [mask] [s-ent] [mask] [rel] [t-ent]  completed in 1795. overview. the [s-ent]  poolbeg[rel] [t-ent]  “peninsula” is home to a number of landmarks including the [s-ent] [mask][rel] [t-ent] , the [s-ent]  pool[mask] lighthouse[rel] [t-ent] , the [s-ent]  irishtown nature park[rel] [t-ent] , the southern part of [s-ent] [mask][rel] [t-ent] ...}
&  \multicolumn{1}{p{2cm}|}{ [ ' connected to land by the',
 ' great south wall',
 ' great south wall',
 'beg',
 ' dublin port',
 "'s main power station,",
 ' structures in',
 '48',
 ' a process to list the',
 ' after the station',
 ', including 3,',
 ' dublin city council',
 ' quarter” on the']} 
&
\raisebox{-\totalheight}{

\includegraphics[width=0.4\textwidth]{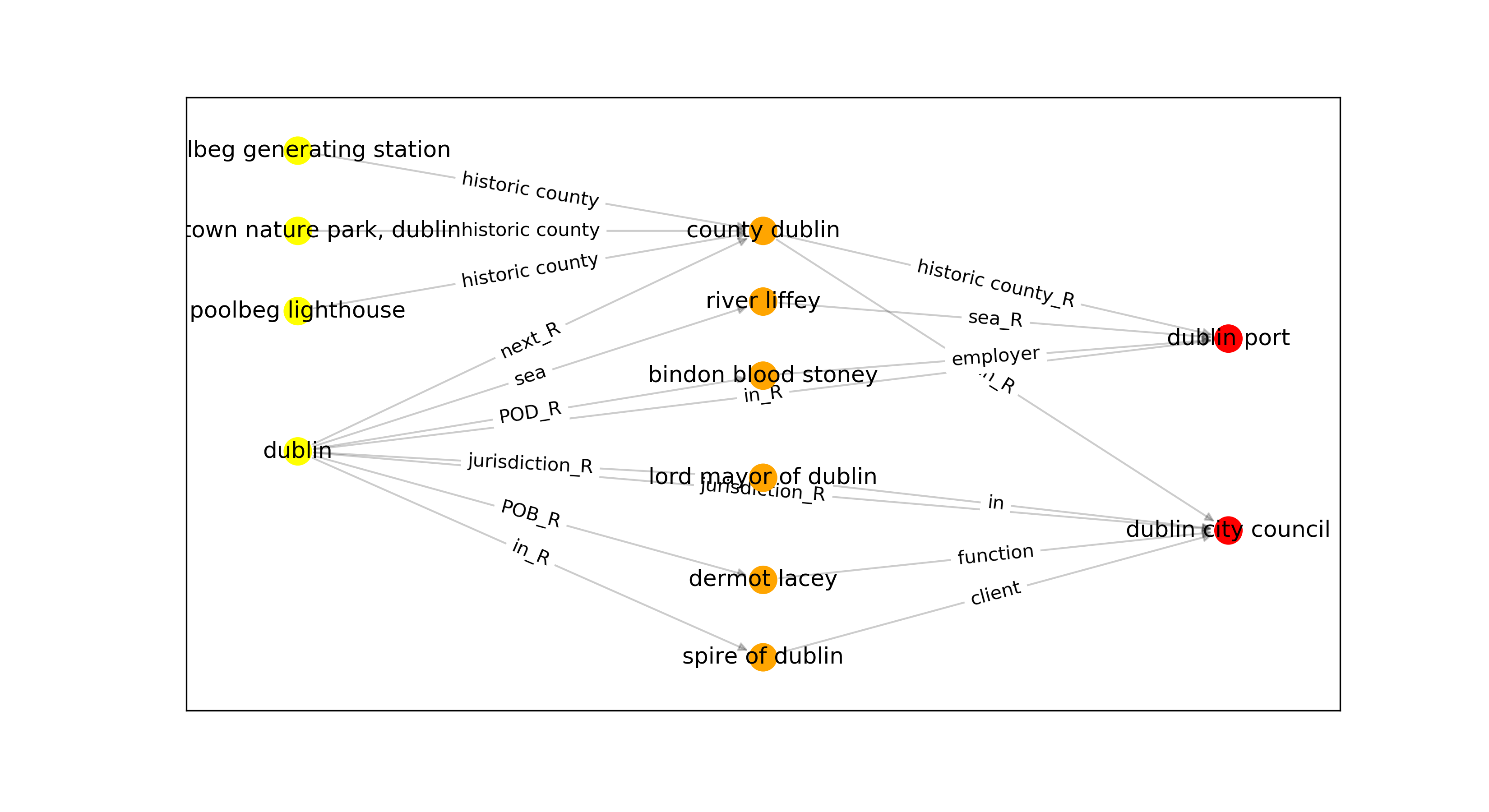}}
&
\raisebox{-\totalheight}{
\includegraphics[width=0.3\textwidth]{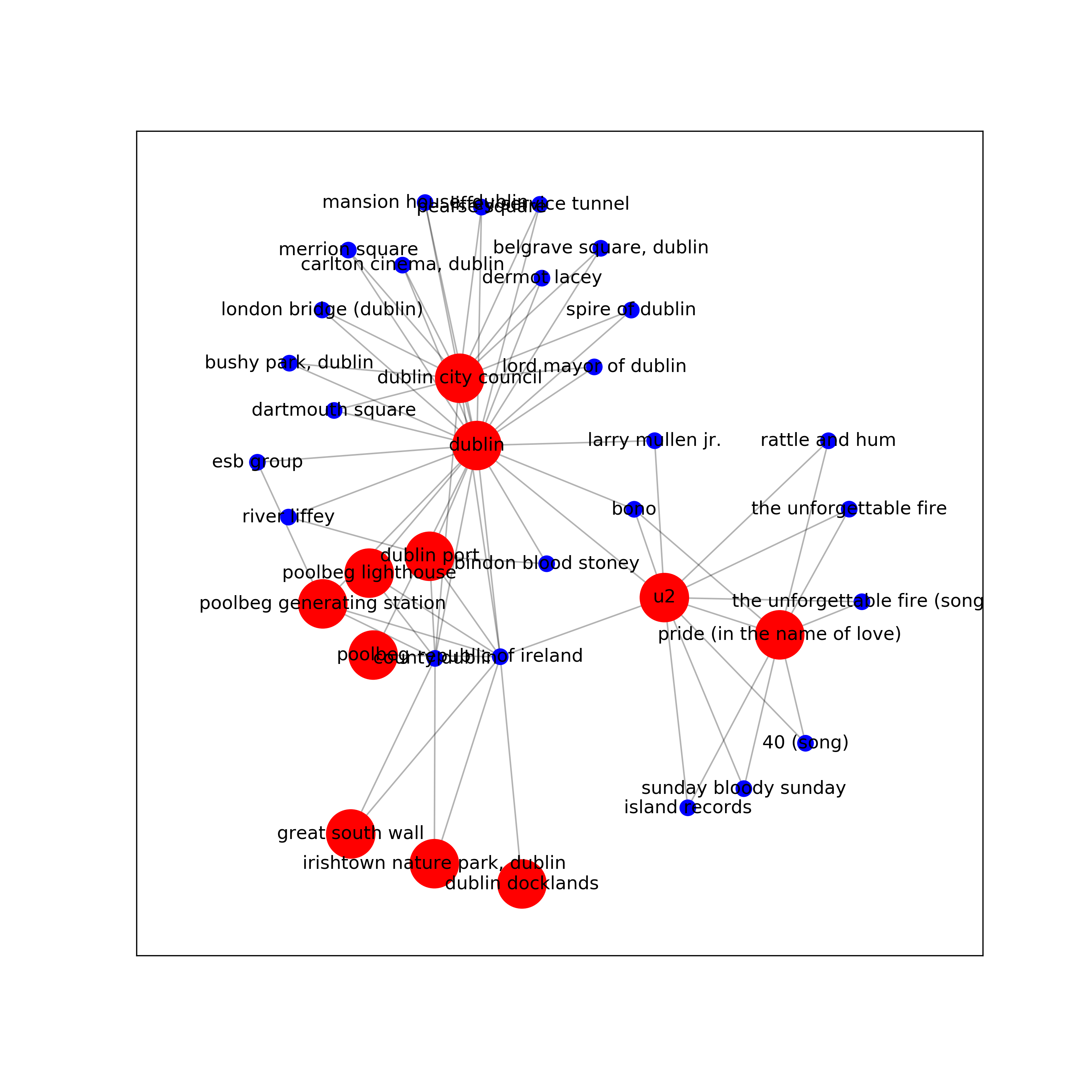}}\\
\midrule
\multicolumn{1}{p{1cm}|}{Rylstone} & \multicolumn{1}{p{2cm}|}{it is situated very near to [s-ent] [mask][rel] [t-ent]  and about 6 miles south west[mask] [s-ent] [mask]ington[rel] [t-ent] . the population of the [s-ent]  civil parish[rel] [t-ent]  as of the 2011 census was 160.  [s-ent] rylstone railway station[rel] [t-ent]  opened in 1902, closed to passengers in 1930, and closed completely in 1969....}
&  \multicolumn{1}{p{2cm}|}{ [' craven',
 ' cracoe',
 ' of',
 ' grass',
 ' the inspiration for',
 ' tour de france',
 'stone',
 ' by will'...]} 
&
\raisebox{-\totalheight}{
\includegraphics[width=0.4\textwidth]{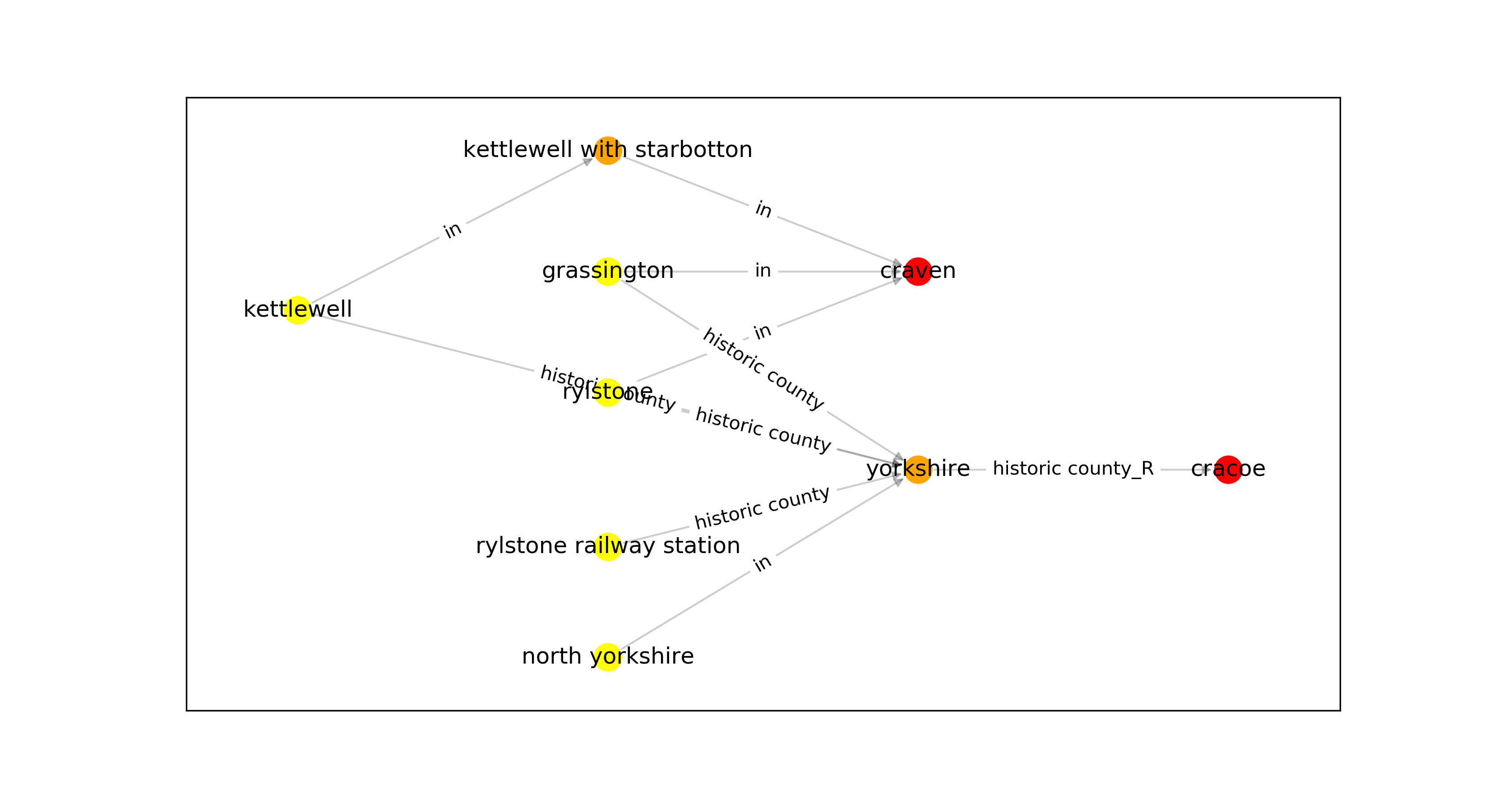}}
&
\raisebox{-\totalheight}{
\includegraphics[width=0.3\textwidth]{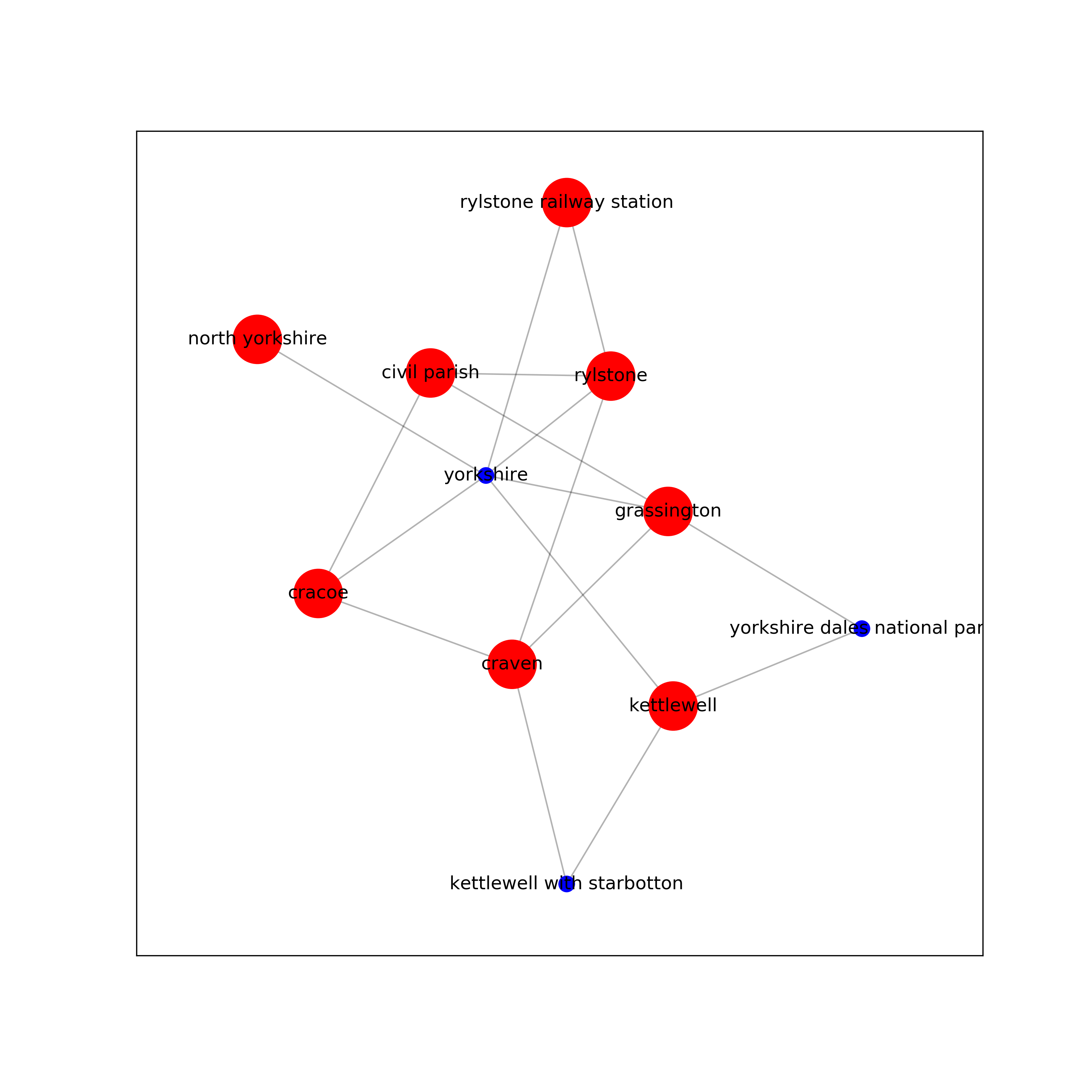}}\\
\midrule
\multicolumn{1}{p{1cm}|}{Karpinsk} & \multicolumn{1}{p{2cm}|}{ologist [s-ent] [mask] [rel] [t-ent] . history.[mask]the settlement of bogoslovsk () was founded in either 1759 or in 1769. it remained one of the largest [s-ent]  copper[rel] [t-ent]  production centers in the [s-ent]  urals[rel] [t-ent] [mask] [s-ent] [mask][rel] [t-ent]  deposits started to be mined in 1911.....}
&  \multicolumn{1}{p{2cm}|}{ [' alexander karpinsky',
 ' until 1917.',
 ' coal',
 'erman civilians, who',
 ' and',
 ' years of',
 ' forest laborers. moreover',
 ' in',
 ' the',
 ' framework of the',
 ' districts',
 ' karpinsk',
 'insk'...]} 
&
\raisebox{-\totalheight}{
\includegraphics[width=0.4\textwidth]{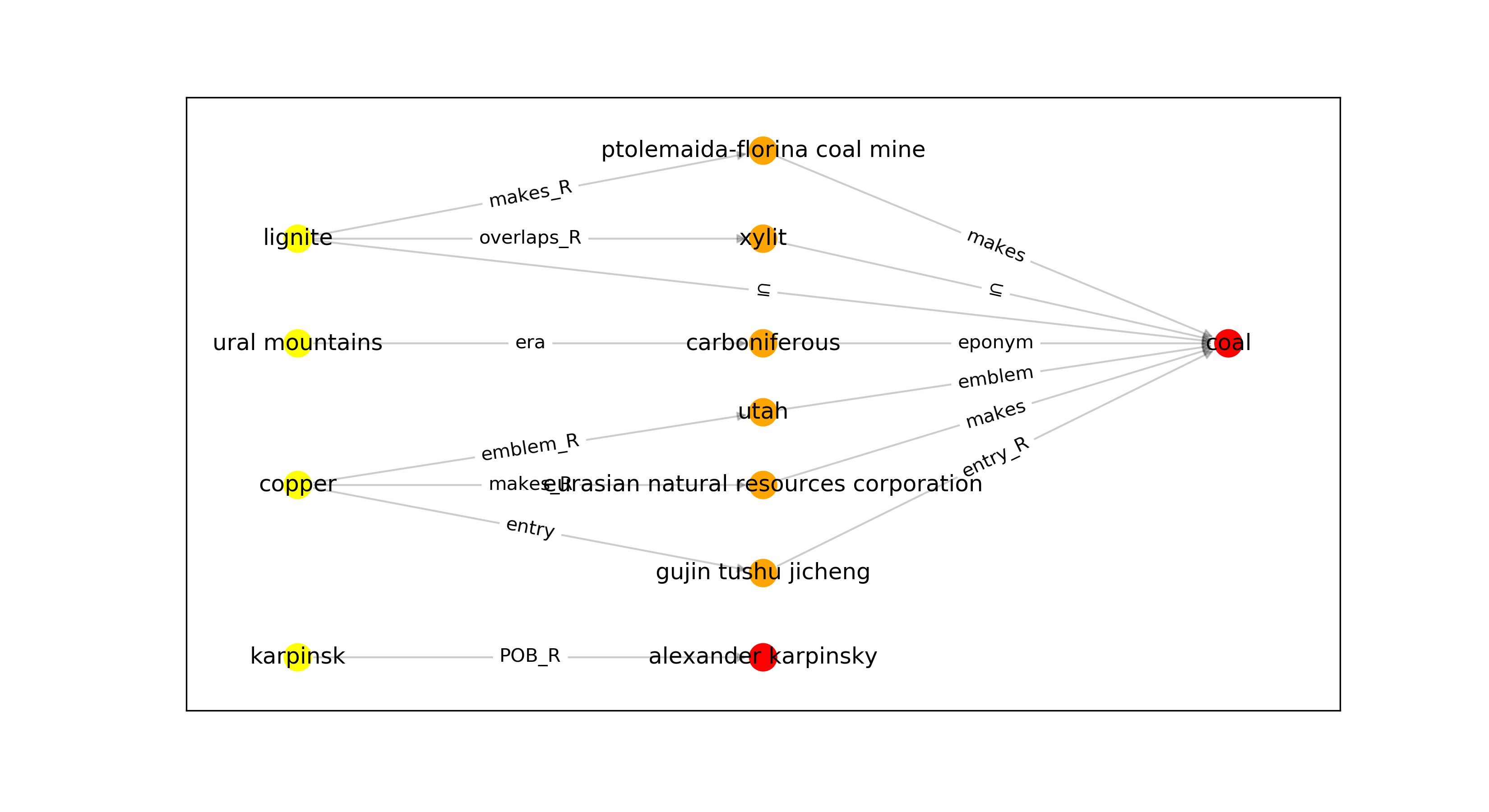}}
&
\raisebox{-\totalheight}{
\includegraphics[width=0.3\textwidth]{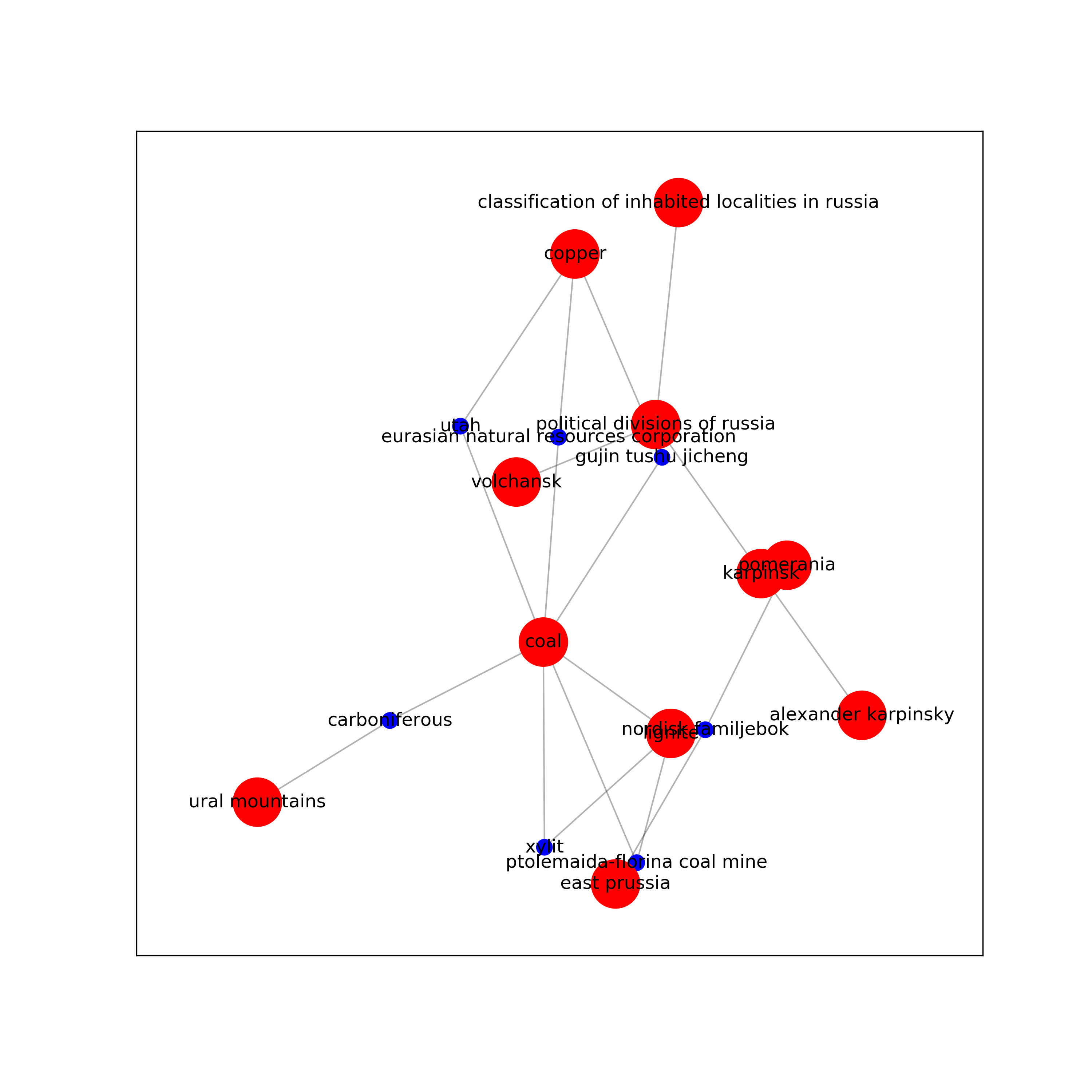}}\\

\midrule
\multicolumn{1}{p{1cm}|}{3 (The X-Files)} & \multicolumn{1}{p{2cm}|}{[s-ent] [mask][mask][rel] [t-ent] ". [s-ent]  gillian anderson[rel] [t-ent]  is absent[mask][mask] episode as she was on leave to give birth to her daughter piper at the time. this episode was the first[mask] not appear. reception. ratings. "3" premiered on the [s-ent]  fox network[rel] [t-ent]  on, and was first broadcast in the [s-ent]  united kingdom[rel] [t-ent].....}
&  \multicolumn{1}{p{2cm}|}{ [ 'ny had',
 ' episode',
 'born again',
 ' from the',
 ' in which scully did',
 '. it was',
 'egall',
 ' metacritic',
 ' as "wretched',
 ' fact that',
 ' background noise for a',
 ' heavy-handed attempts at',
 ' glen morgan',
 ' doing an episode on']} 
&
\raisebox{-\totalheight}{
\includegraphics[width=0.4\textwidth]{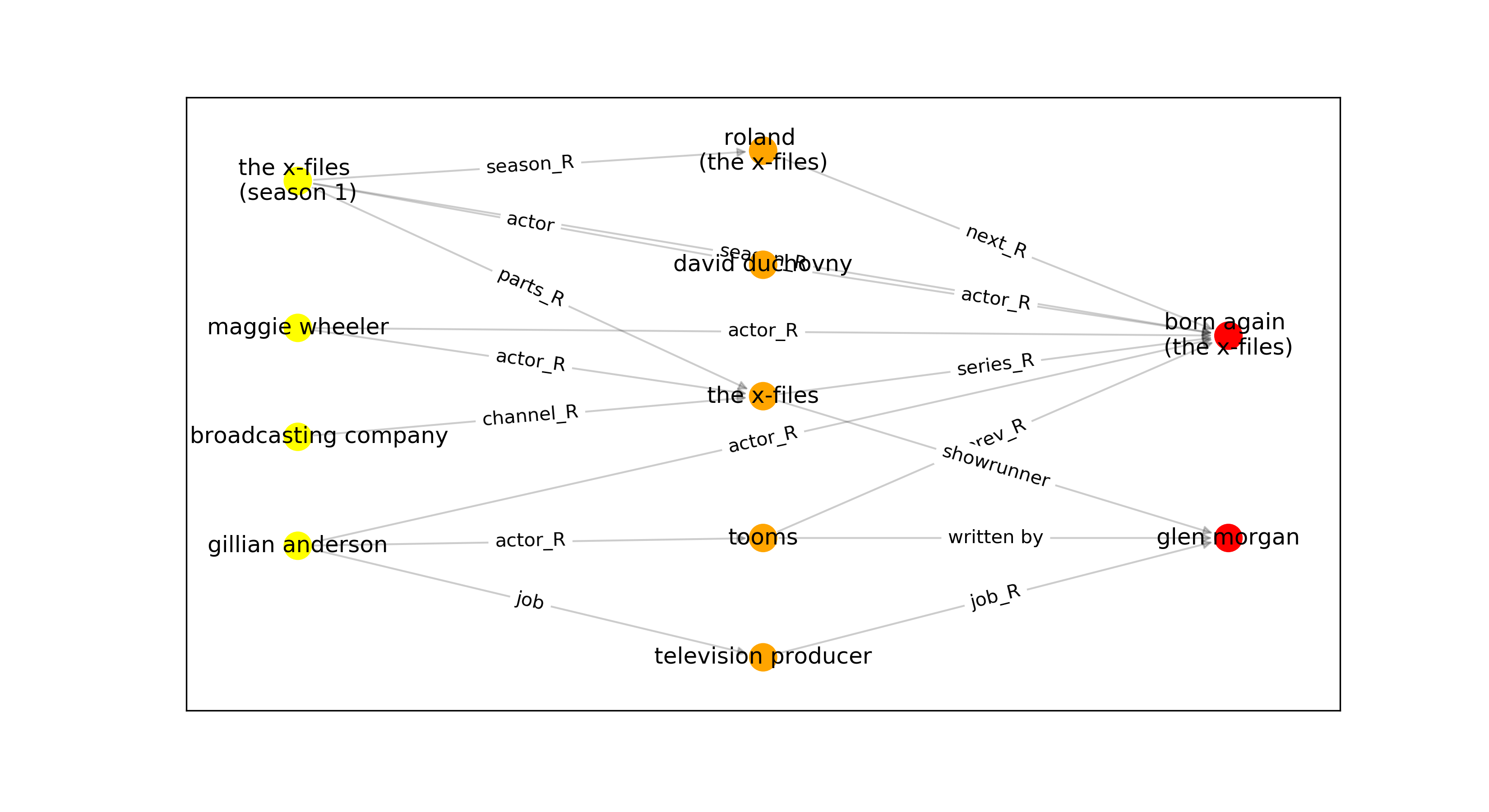}}
&
\raisebox{-\totalheight}{
\includegraphics[width=0.3\textwidth]{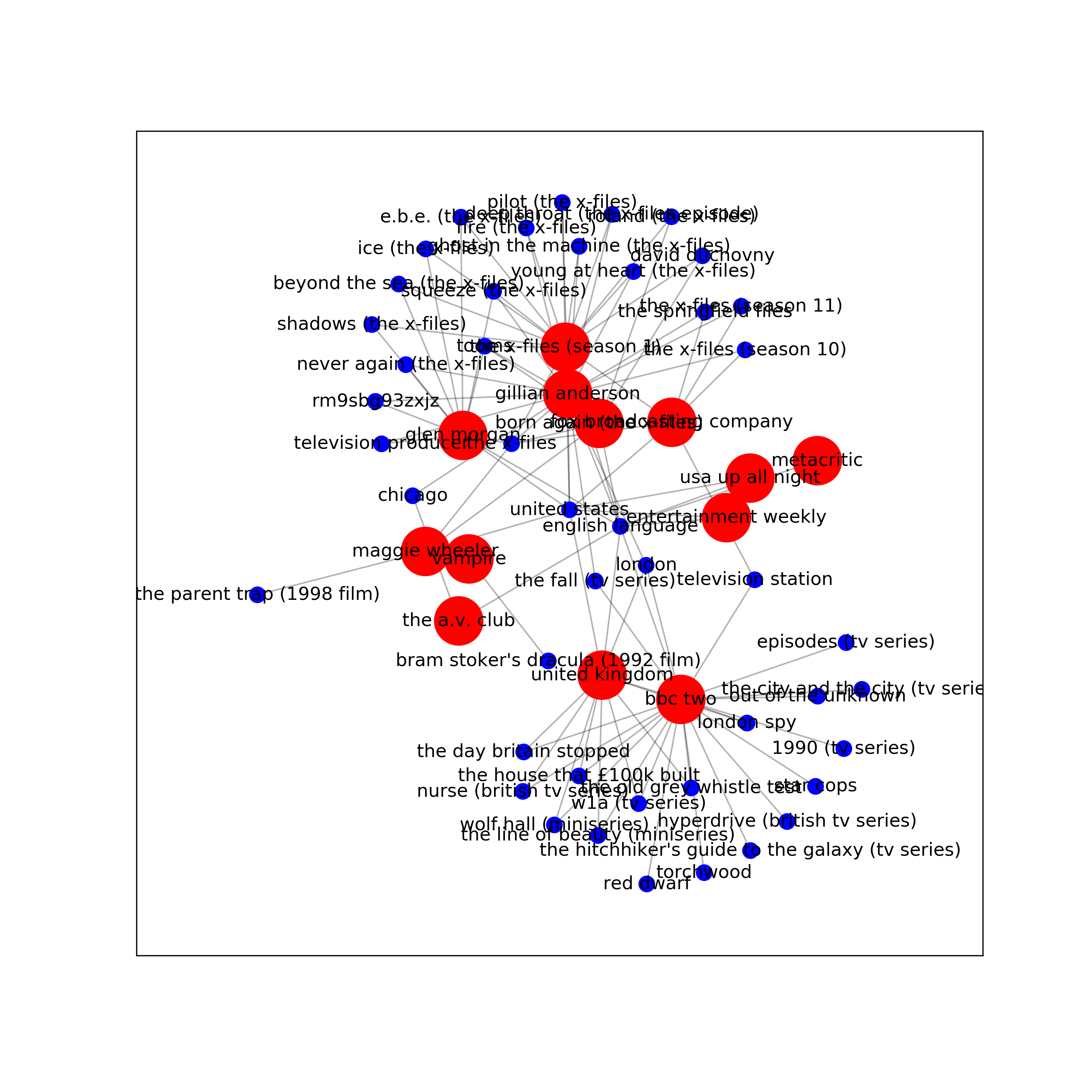}}\\
\bottomrule
\end{tabular}
\caption{Example of Pre-training data points (Part 1).}
\label{tab:pretrain1}
\end{table*}

\begin{table*}[h]
\scriptsize
\begin{tabular}[t]{c|c|c|c|c} \toprule
\textbf{Title}   & \textbf{Masked Text}  & \textbf{Ground Truth} & \textbf{Dependency Graph} & \textbf{K-Hop Graph}  \\ \midrule

\multicolumn{1}{p{1cm}|}{Shen Chun-shan} & \multicolumn{1}{p{2cm}|}{his memoirs, he suffered his second stroke[mask][mask], even after his second stroke, he continued writing; his series of biographies of five go masters [s-ent] [mask][mask][mask][rel] [t-ent] , [s-ent]  minoru kit[mask][rel] [t-ent] .....}
&  \multicolumn{1}{p{2cm}|}{ ['. however',
 ' go seigen',
 'ani',
 ' 2007, he',
 ' was hospital',
 ' hsinchu',
 'after surgery',
 ' scale',
 ' continuing to improve.',
 ' his coma. in'...]} 
&
\raisebox{-\totalheight}{
\includegraphics[width=0.4\textwidth]{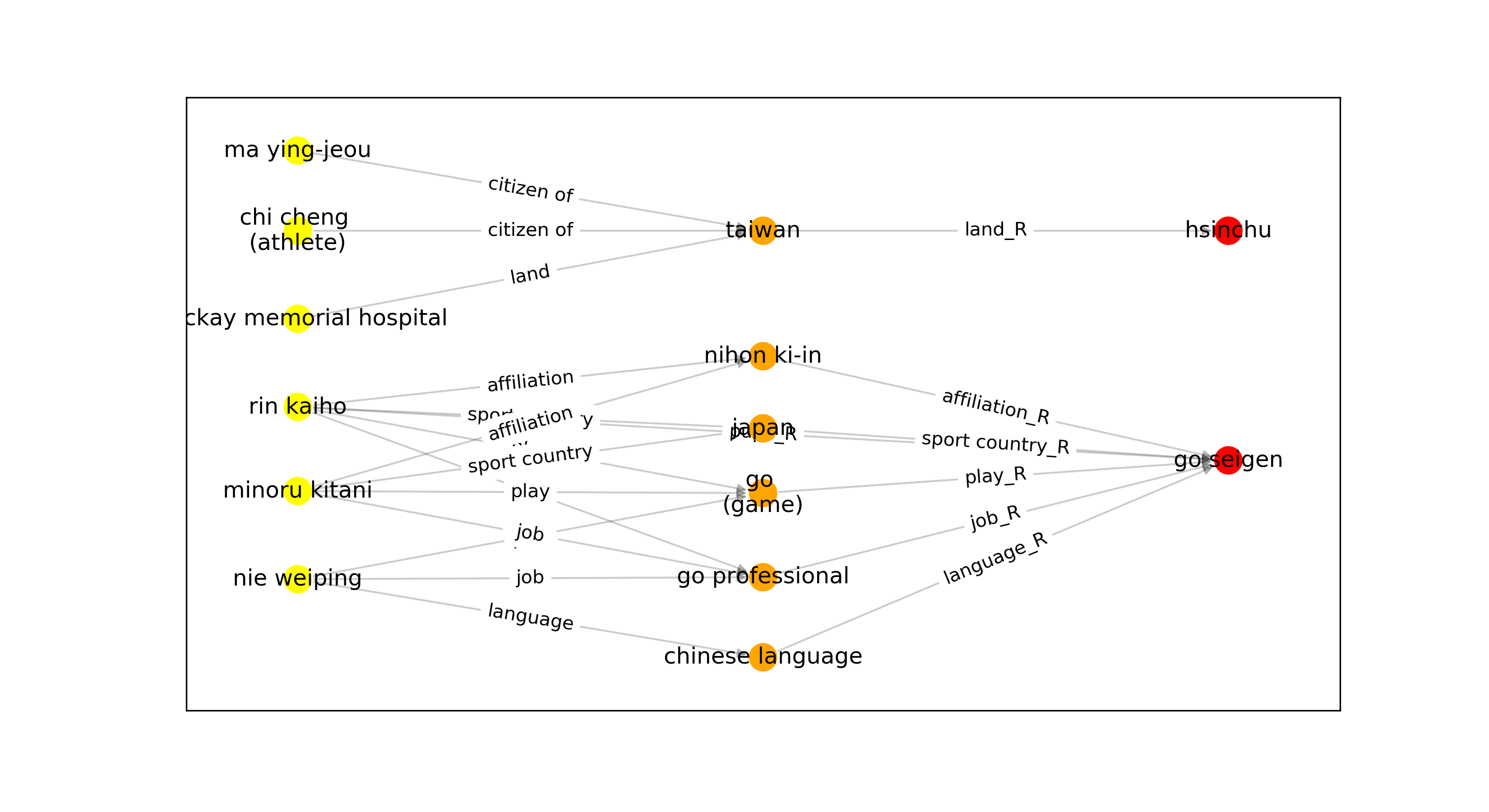}}
&
\raisebox{-\totalheight}{
\includegraphics[width=0.3\textwidth]{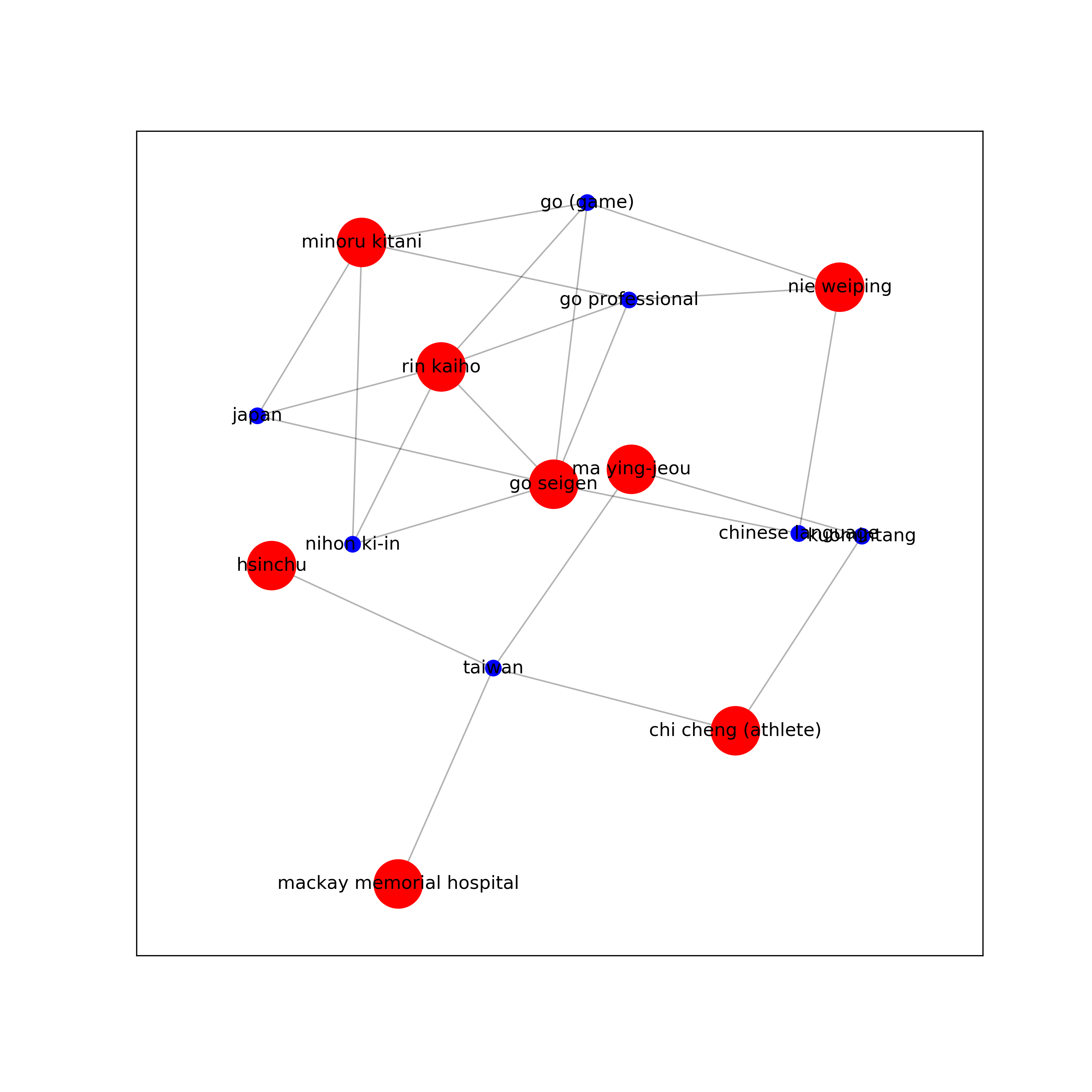}}\\

\midrule
\multicolumn{1}{p{1cm}|}{2007 Florida Gators football team} & \multicolumn{1}{p{2cm}|}{[s-ent]  tim[mask][mask][rel] [t-ent]  completed 22 of 27 passes for 281 yards passing and also ran for[mask] yards on 6 carries. [s-ent] [mask] [rel] [t-ent]  carried the ball 11 times for 113 yards[mask] two touchdowns and also caught 9 passes for 110[mask] receiving, becoming the first player in school history .....}
&  \multicolumn{1}{p{2cm}|}{ [' tebow',
 ' 35',
 ' percy harvin',
 ' and',
 ' yards',
 ' 30–9',
 ' renewed their budding',
 ' gamecocks',
 'gator',
 ' quarterback',
 ' set a career-high',
 ' of these five rushing',
 '.',
 ' percy harvin',
 ' sinus infection.',
 'ators',
 ' touchdown']} 
&
\raisebox{-\totalheight}{
\includegraphics[width=0.4\textwidth]{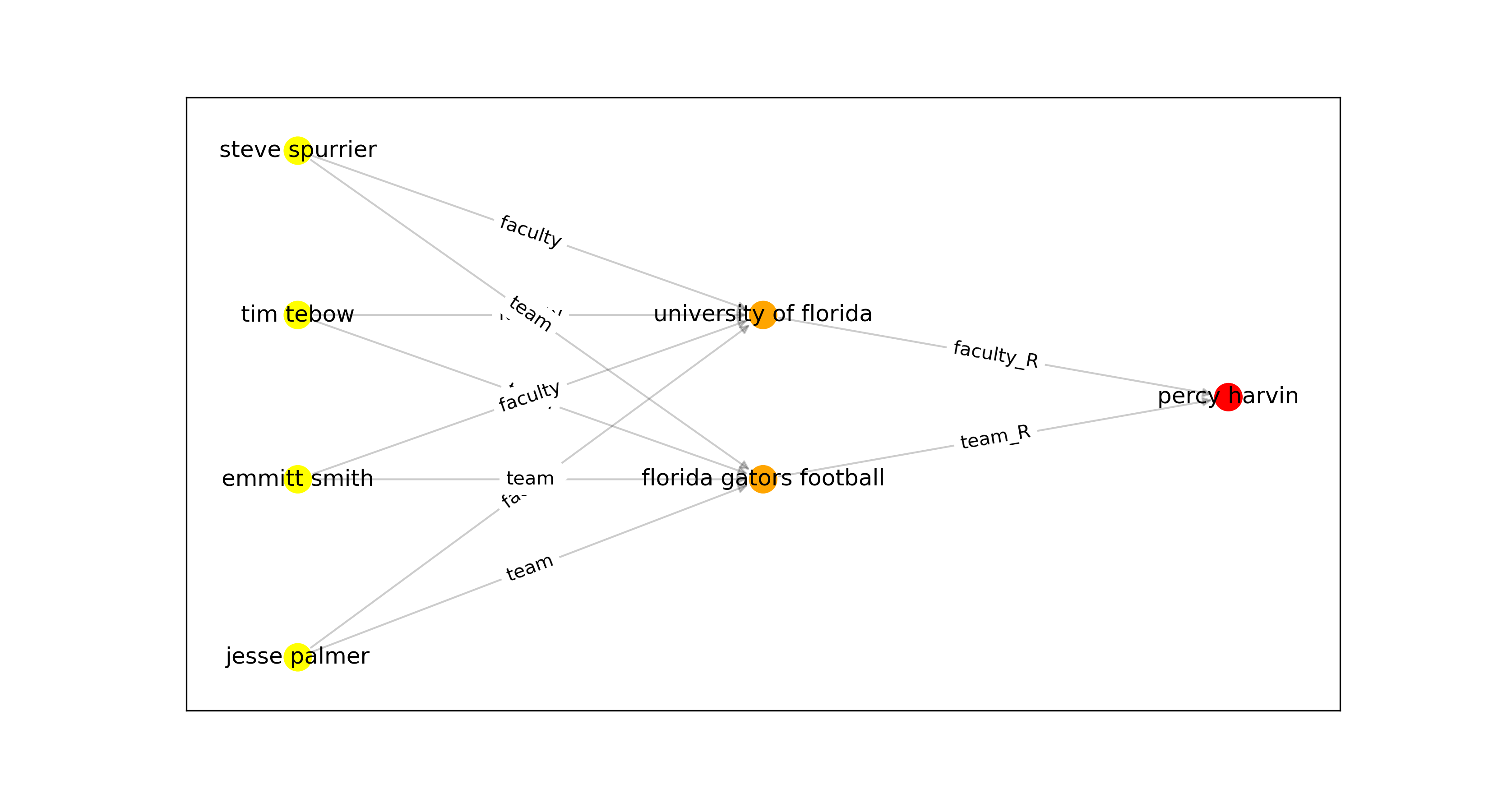}}
&
\raisebox{-\totalheight}{
\includegraphics[width=0.3\textwidth]{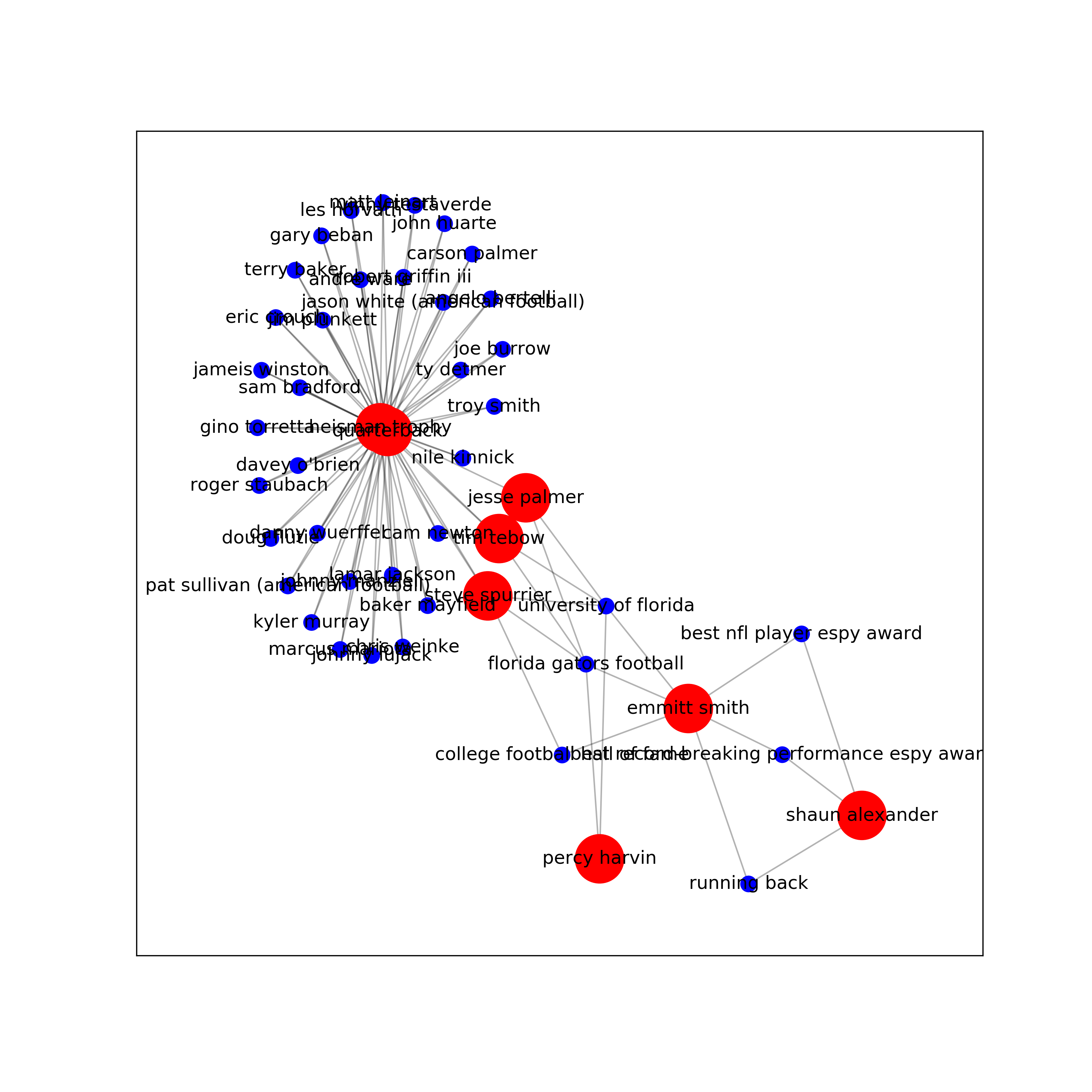}}\\

\midrule
\multicolumn{1}{p{1cm}|}{Judgment Day (Awesome Comics)} & \multicolumn{1}{p{2cm}|}{[s-ent]  alan moore[rel] [t-ent]  used "judgment day" to reject the violent, deconstructive clichés of 1990s comics inadvertently caused by his own work on " [s-ent] watchmen[rel] [t-ent] ", "" and " [s-ent] saga of the[mask][mask][rel] [t-ent] " and uphold the values of classic superhero comics. the series deals with a metacommentary of the notion of retcons to super-hero histories as [s-ent]  alan moore[rel] [t-ent] [mask]  for the characters of [s-ent] [mask][mask][rel] [t-ent] , to replace the shared universe they left when [s-ent]  rob liefeld[rel] [t-ent]  left image several years earlier. plot. in[mask], mick tombs/ [s-ent] knightsabre[rel] [t-ent].....}
&  \multicolumn{1}{p{2cm}|}{ [ ' swamp thing',
 ' himself creates a new backstory',
 ' awesome comics',
 ' 1997',
 'riptide',
 ' knightsabre appears to be',
 ' and sw',
 ' badrock',
 ' supreme',
 'by',
 ' analyzing',
 ' cybernetic young',
 ' it, and it has',
 'ue out',
 ', administrator for youngblood']} 
&
\raisebox{-\totalheight}{
\includegraphics[width=0.4\textwidth]{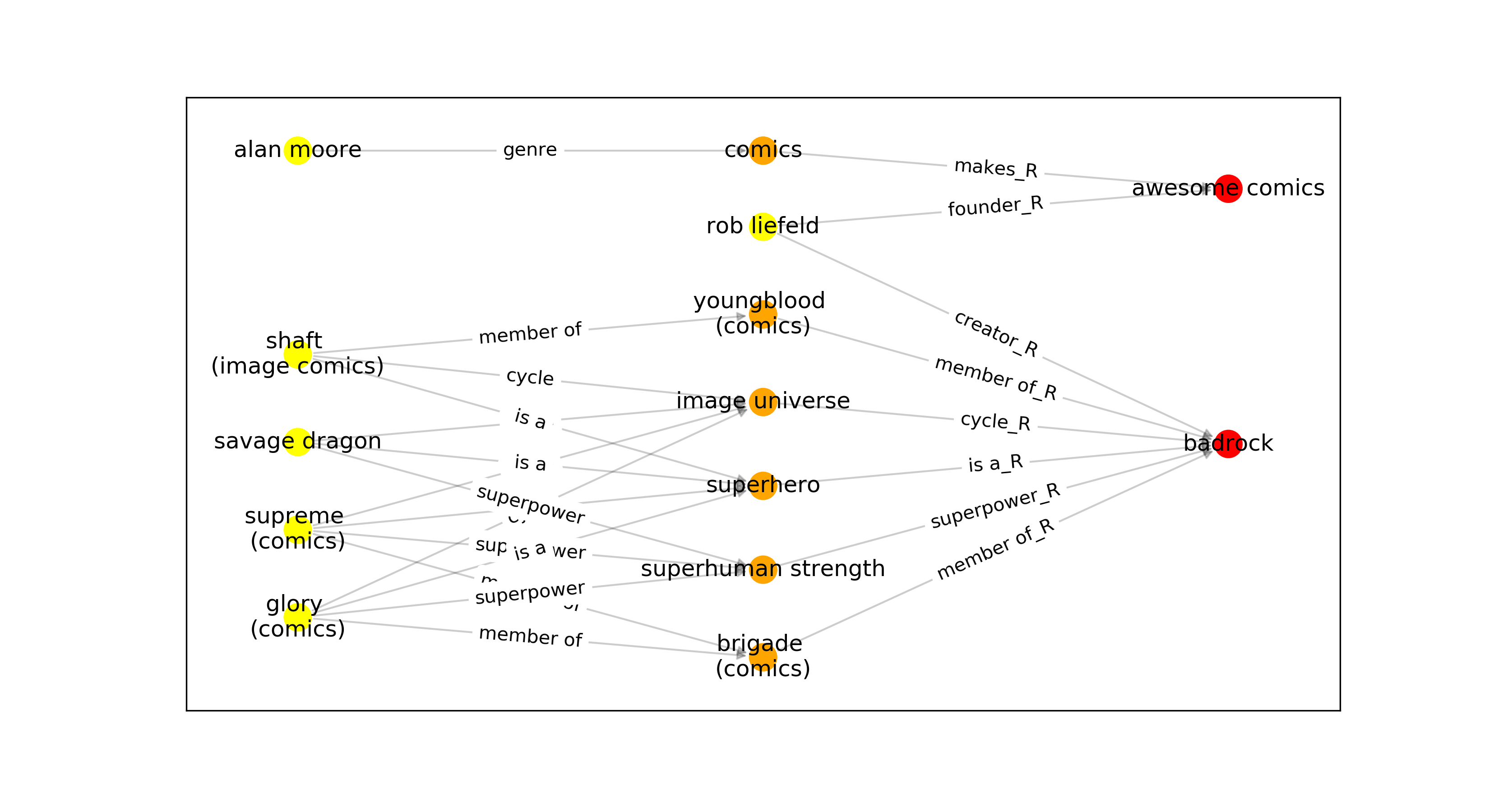}}
&
\raisebox{-\totalheight}{
\includegraphics[width=0.3\textwidth]{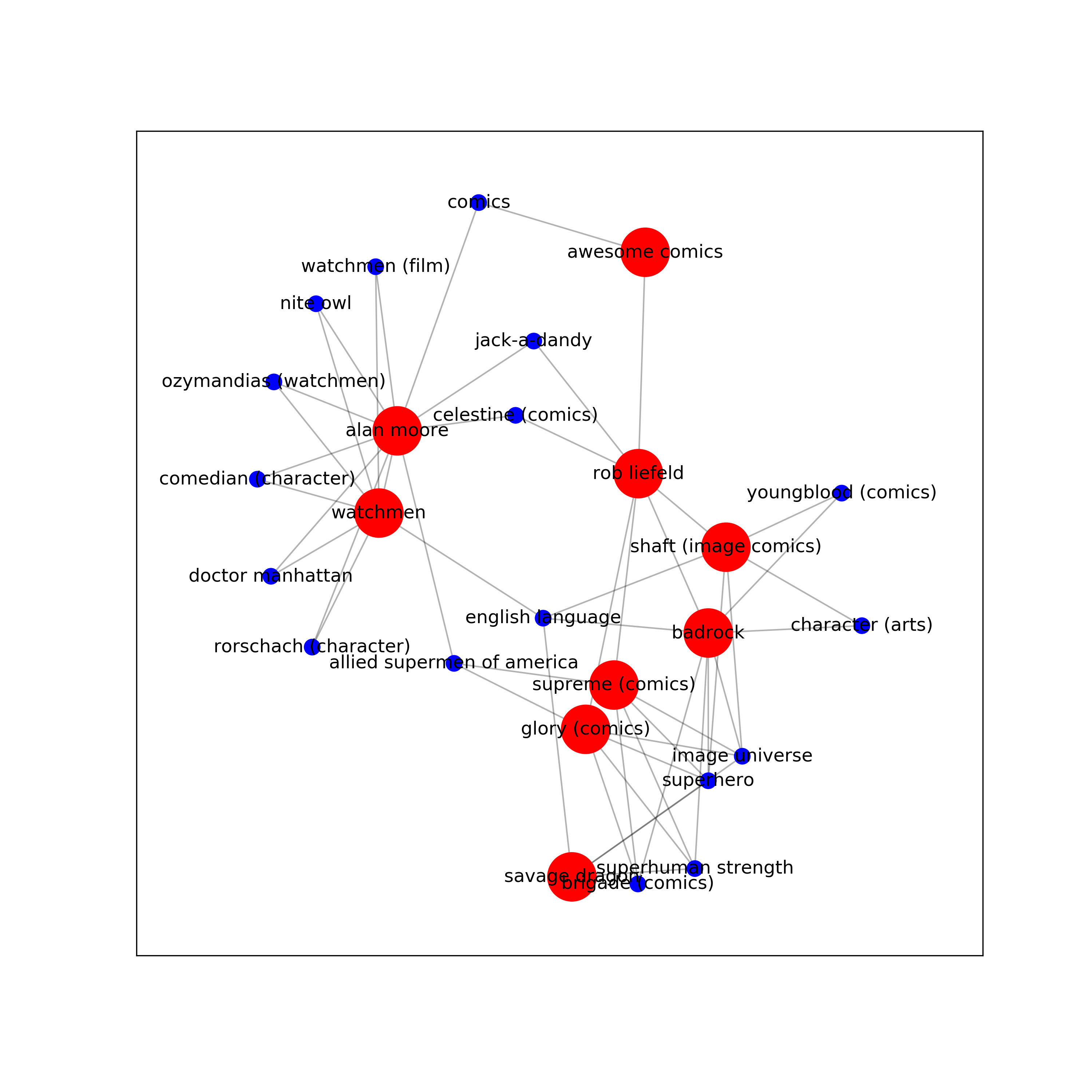}}\\

\bottomrule
\end{tabular}
\caption{Example of Pre-training data points (Part 2).}
\label{tab:pretrain}
\end{table*}

\begin{table*}[ht]
\centering
\hspace{-.5in}
\begin{tabular}{c|c|l} \toprule
\textbf{Question}   & \textbf{Answer}  & \textbf{Reasoning Paths as Rationale}  \\ \midrule
\multicolumn{1}{p{4cm}|}{ southern soul was considered the sound of what independent record label } & \multicolumn{1}{p{2cm}|}{['Motown']} & \multicolumn{1}{p{12cm}|}{$ \text{ soul music } \xrightarrow{ \text{genre-R} } \ ?\  \xrightarrow{ \text{label} } \ ?\  $}\\ 
 & & $ \text{ independent record label } \xrightarrow{ \text{belong} } \ ?\  \xrightarrow{ \text{is a-R} } \ ?\  $\\ \midrule
\multicolumn{1}{p{4cm}|}{ who is the bad guy in lord of the rings } & \multicolumn{1}{p{2cm}|}{['Sauron']} & \multicolumn{1}{p{12cm}|}{$ \text{ the lord of the rings (film series) } \xrightarrow{ \text{theme} } \ ?\  \xrightarrow{ \text{characters} } \ ?\  $}\\ \midrule
\multicolumn{1}{p{4cm}|}{ where was the mona lisa kept during ww2 } & \multicolumn{1}{p{2cm}|}{['the Ingres Museum', "Château d'Amboise", 'Château de Chambord', 'the Loc - Dieu Abbey']} & \multicolumn{1}{p{12cm}|}{$ \text{ mona lisa } \xrightarrow{ \text{creator} } \ ?\  \xrightarrow{ \text{tomb} } \ ?\  $}\\ 
& & $ \text{  world war 2 } \xrightarrow{ \text{take place} } \ ?\  \xrightarrow{ \text{located-R} } \ ?\  $ \\ \midrule
\multicolumn{1}{p{4cm}|}{ who have won the world cup the most times } & \multicolumn{1}{p{2cm}|}{['Brazil']} & \multicolumn{1}{p{12cm}|}{$ \text{ fifa world cup } \xrightarrow{ \text{parts} } \ ?\  \xrightarrow{ \text{land} } \ ?\  $}\\ \midrule
\multicolumn{1}{p{4cm}|}{ who wrote the song the beat goes on } & \multicolumn{1}{p{2cm}|}{['Sonny Bono']} & \multicolumn{1}{p{12cm}|}{$ \text{ song } \xrightarrow{ \text{album type-R} } \ ?\  \xrightarrow{ \text{author} } \ ?\  $}\\ \midrule
\multicolumn{1}{p{4cm}|}{ who plays mrs. potato head in toy story } & \multicolumn{1}{p{2cm}|}{['Estelle Harris']} & \multicolumn{1}{p{12cm}|}{$ \text{ toy story } \xrightarrow{ \text{series} } \ ?\  \xrightarrow{ \text{VO} } \ ?\  $}\\ \midrule
\multicolumn{1}{p{4cm}|}{ who plays caroline on the bold and beautiful } & \multicolumn{1}{p{2cm}|}{['Linsey Godfrey']} & \multicolumn{1}{p{12cm}|}{$ \text{ the bold and the beautiful } \xrightarrow{ \text{in work-R} } \ ?\  \xrightarrow{ \text{actor} } \ ?\  $}\\ \midrule
\multicolumn{1}{p{4cm}|}{ where are the fruits of the spirit found in the bible } & \multicolumn{1}{p{2cm}|}{['Epistle to the Galatians']} & \multicolumn{1}{p{12cm}|}{$ \text{ bible } \xrightarrow{ \text{parts} } \ ?\  \xrightarrow{ \text{parts} } \ ?\  $}\\ \midrule
\multicolumn{1}{p{4cm}|}{ who is the only kaurava who survived the kurukshetra war } & \multicolumn{1}{p{2cm}|}{['Yuyutsu']} & \multicolumn{1}{p{12cm}|}{$ \text{ kaurava } \xrightarrow{ \text{in work} } \ ?\  \xrightarrow{ \text{in work-R} } \ ?\  $}\\
& & $\text{Kurukshetra War} \xrightarrow{ \text{location} } \xrightarrow{ \text{live in-R} } $ \\
\midrule
\multicolumn{1}{p{4cm}|}{ what is the deepest depth in the oceans } & \multicolumn{1}{p{2cm}|}{['Mariana Trench']} & \multicolumn{1}{p{12cm}|}{$ \text{ ocean } \xrightarrow{ \text{in} } \ ?\  \xrightarrow{ \text{lowest point} } \ ?\  $}\\ \midrule
\multicolumn{1}{p{4cm}|}{ where did the french national anthem come from } & \multicolumn{1}{p{2cm}|}{['Strasbourg']} & \multicolumn{1}{p{12cm}|}{$ \text{ national anthem } \xrightarrow{ \text{is a-R} } \ ?\  \xrightarrow{ \text{released in} } \ ?\  $}\\   \bottomrule
\end{tabular}
\caption{Example of QA prediction with reasoning path on NQ (part 1).}
\label{tab:explain1}
\end{table*}

\begin{table*}[ht]
\centering
\hspace{-.5in}
\begin{tabular}{c|c|l} \toprule
\textbf{Question}   & \textbf{Answer}  & \textbf{Generated Reasoning Paths as Rationale}  \\  \midrule
\multicolumn{1}{p{4cm}|}{ who sings the song where have all the flowers gone } & \multicolumn{1}{p{2cm}|}{['Pete Seeger']} & \multicolumn{1}{p{12cm}|}{$ \text{ song } \xrightarrow{ \text{album type-R} } \ ?\  \xrightarrow{ \text{actor} } \ ?\  $}\\ \midrule
\multicolumn{1}{p{4cm}|}{ who discovered some islands in the bahamas in 1492 } & \multicolumn{1}{p{2cm}|}{['Christopher Columbus']} & \multicolumn{1}{p{12cm}|}{$ \text{ the bahamas } \xrightarrow{ \text{entry} } \ ?\  \xrightarrow{ \text{entry-R} } \ ?\  $}\\ \midrule
\multicolumn{1}{p{4cm}|}{ which type of wave requires a medium for transmission } & \multicolumn{1}{p{2cm}|}{['mechanical waves', 'heat energy', 'Sound']} & \multicolumn{1}{p{12cm}|}{$ \text{ wave } \xrightarrow{ \text{belong-R} } \ ?\  \xrightarrow{ \text{belong-R} } \ ?\  $}\\ \midrule
\multicolumn{1}{p{4cm}|}{ land conversion through burning of biomass releases which gas } & \multicolumn{1}{p{2cm}|}{['traces of methane', 'carbon monoxide', 'hydrogen']} & \multicolumn{1}{p{12cm}|}{$ \text{ gas } \xrightarrow{ \text{belong-R} } \ ?\  \xrightarrow{ \text{as-R} } \ ?\  $}\\ \midrule
\multicolumn{1}{p{4cm}|}{ the sum of the kinetic and potential energies of all particles in the system is called the } & \multicolumn{1}{p{2cm}|}{['internal energy']} & \multicolumn{1}{p{12cm}|}{$ \text{ kinetic energy } \xrightarrow{ \text{belong} } \ ?\  \xrightarrow{ \text{belong-R} } \ ?\  $}\\ 
 & & $ \text{ potential energy } \xrightarrow{ \text{belong} } \ ?\  \xrightarrow{ \text{belong-R} } \ ?\  $\\ \midrule
\multicolumn{1}{p{4cm}|}{ who did seattle beat in the super bowl } & \multicolumn{1}{p{2cm}|}{['Denver Broncos']} & \multicolumn{1}{p{12cm}|}{$ \text{ super bowl } \xrightarrow{ \text{organizer} } \ ?\  \xrightarrow{ \text{league-R} } \ ?\  $}\\ \midrule
\multicolumn{1}{p{4cm}|}{ what is the name of the girl romeo loved before juliet } & \multicolumn{1}{p{2cm}|}{['Rosaline']} & \multicolumn{1}{p{12cm}|}{$ \text{ romeo } \xrightarrow{ \text{in work} } \ ?\  \xrightarrow{ \text{in work-R} } \ ?\  $}\\ \midrule
\multicolumn{1}{p{4cm}|}{ who will get relegated from the premier league 2016/17 } & \multicolumn{1}{p{2cm}|}{[ 'Hull City', 'Sunderland', 'Middlesbrough']} & \multicolumn{1}{p{12cm}|}{$ \text{ premier league } \xrightarrow{ \text{league-R} } \ ?\  \xrightarrow{ \text{POB} } \ ?\  $}\\ \midrule
\multicolumn{1}{p{4cm}|}{ actress in the girl with the dragon tattoo swedish } & \multicolumn{1}{p{2cm}|}{['Noomi Rapace']} & \multicolumn{1}{p{12cm}|}{$ \text{ sweden } \xrightarrow{ \text{speaking} } \ ?\  \xrightarrow{ \text{mother tongue-R} } \ ?\  $}\\
\bottomrule
\end{tabular}
\caption{Example of QA prediction with reasoning path on NQ (part 2).}
\label{tab:explain2}
\end{table*}

\section{Dataset Details}\label{sec:exp_detail}

Below shows details for each dataset, and the detailed dataset split is shown in Figure~\ref{tab:data}

\paragraph{Natural Questions}\cite{DBLP:journals/tacl/KwiatkowskiPRCP19} contains questions from Google search queries, and the answers are text spans in Wikipedia. We report short answer Exact Match (EM) performance. The open version of this dataset is obtained by discarding answers with more than 5 tokens.

\paragraph{WebQuestions (WQ)} \cite{DBLP:conf/emnlp/BerantCFL13} contains questions from Google Suggest API, and the answers are entities in Freebase.

\paragraph{TriviaQA} \cite{DBLP:conf/acl/JoshiCWZ17} contains trivia questions and answers are text spans from the Web. We report Exact Match (EM) performance. We use its unfiltered version for evaluation.

\paragraph{HotpotQA} \cite{DBLP:conf/emnlp/Yang0ZBCSM18} is a multi-hop QA dataset. There are two evaluation settings. In the \textit{distractor setting}, 10 candidate paragraphs are provided for each question, of which there are two golden paragraphs. In the \textit{full-wiki setting}, a model is required to extract paragraphs from the entire Wikipedia. We report Exact Match (EM) on full-wiki setting.

\paragraph{Complex WebQuestions}\cite{DBLP:conf/naacl/TalmorB18} is a dataset that composite simple one-hot questions in WebQuestionsSP by extending entities or adding constraints, so that each question eequires complex reasoning to solve.

\paragraph{WebQuestionsSP}\cite{DBLP:conf/acl/YihCHG15} is annotated dataset from WebQuestions, such taht each quetsion is answerable using Freebase via a SQL query.

\section{Discussion with Previous Works}

\paragraph{Compare with FILM}

Though FILM has the advantage of end-to-end training and easily modification of knowledge memory, it simply stacks $\mathcal{KG}$ module on top of \texttt{LM} without interaction, and can only handle one-hop relational query that is answerable by $\mathcal{KG}$. 
Our approach, \method, follows the same \emph{memory} idea by encoding $\mathcal{KG}$ into \texttt{LM} parameter, and we desire \texttt{LM} and $\mathcal{KG}$ reasoning module could interact and collaboratively improve each other.

Notably, \method with $T=1$ shares a similar design with FILM. The major differences are: 1) they store every triplet as a key-value pair,
while we explicitly keep the $\mathcal{KG}$ adjacency matrix and conduct a random walk, which has smaller search space and is more controllable. 2) They add the memory on top of \texttt{LM}, and thus the knowledge could not help language understanding, and FILM could mainly help wikipedia-answerable questions. Instead, we insert the \texttt{KIL} layer amid \texttt{LM} layers to encourage interaction, and thus the model could also benefit encoder-decoder model (as shown above).

\paragraph{Compare with Previous Path-Based Reasoning and Retrieval Pre-Training}

Note that as our definition of entity state $\pib_i$ and relation action $\rb_i$ are both continuous probabilistic vector, the whole  $\mathcal{KG}$ Reasoning is fully differentiable and thus could be integrated into \texttt{LM} seamlessly and trained end-to-end. This is different from previous path traversal works such as DeepPath~\cite{DBLP:conf/emnlp/XiongHW17} and MINERVA~\cite{DBLP:conf/iclr/DasDZVDKSM18}, which defines state and action as discrete and could only be trained via reinforcement learning rewards. The reasoner training is also different from passage retrieval pre-training~\cite{DBLP:journals/corr/abs-2002-08909, DBLP:conf/ACL/SachanPSKPHC20}, as the passage are naturally consisted of discrete tokens, and thus the reader is still required to re-encode the question with each passage, and different objectives are required to train retriever and reader separately.

\paragraph{Discussion of Graph Walking-based Reasoning vs Graph Neural Networks}~\label{sec:gnn}
Recently, Graph Neural Networks (GNNs) have shown superior performance for structured representation learning. There's also a lot of works trying to use GNNs for Question Answering~\citep{DBLP:conf/naacl/YasunagaRBLL21, DBLP:journals/corr/abs-2201-08860}. The one that has very similar motivation with us is GreaseLM. Therefore, a natural question is, whether could we use GNN instead of the non-parametric random walk module, for ODQA?

To answer this question, let's consider a simplest setup of GNN. We could identify initial entities, connected them via a k-hop subgraph, and encode graph with text~\citep{DBLP:journals/corr/abs-2201-08860} or independently~\citep{DBLP:journals/corr/abs-2010-00796}. When we want to retrieve knowledge from graph to LM, normally we just take the contextualized node embedding as input for knowledge fusion.

In this setup, say the answer is $K$-hop away from an initial entity, the ground-truth reasoning path is $e_0, r_1, e_1, r_2, ..., e_{k-1}, r_k, e_{k}=a$. Using our method, we first predict $r1$, transit to $e_1$, and step by step conduct reasoning via walking. However, if we use GNN's final embedding, it requires to pass information from neighbor to itself. Therefore, suppose we have a $K$-layer GNN, the first step should be identify $r_k$, and pass information from answer $e_{k}=a$ to $e_{k-1}$. This is conter-intuitive as we normally cannot assume to know the answer, nor knowling the last step to reach the answer. In situations where all candidate answer is given, like CommonSenseQA, where GreaseLM mainly works on, this problem is less harmful as it's guaranteed to contain the answer in a restricted small graph. However, in open-domain setup, we need to try best to narrow down the search space by following the forward reasoning instead of the backward manner. Therefore, in this work we adopt walking-based reasoning.

\section{Illustration of Pre-Trained Data and Reasoning Paths}

The pre-training samples and reasoning paths (generated by T5-large on NQ dataset) is shown from Table~\ref{tab:pretrain1}-\ref{tab:explain2}.